\documentclass{article}

\usepackage{microtype}
\usepackage{graphicx}
\usepackage{subfigure}
\usepackage{booktabs} 

\usepackage{hyperref}

\usepackage{color,soul}
\usepackage{graphicx}
\usepackage{amsmath}
\usepackage{booktabs}
\usepackage{multirow, makecell}
\usepackage{tikz}
\usepackage{graphicx}
\usepackage{array}
\usepackage{pgfplots}

\usepackage{textcomp}
\usepackage{xcolor}
\usepackage{times}
\usepackage{latexsym}
\usepackage{amsmath}
\usepackage{amsfonts}
\usepackage{lipsum}
\usepackage{multirow}
\usepackage{framed}
\usepackage{graphicx}
\usepackage{pifont}
\usepackage{longtable}
\usepackage{booktabs}
\usepackage{enumitem}
\usepackage{color,soul}

\usepackage[accepted]{icml2020}

\newcommand{\roberta}{$\textsc{RoBERTa}$}

\newcommand{\eric}[1]{}
\newcommand{\zhuohan}[1]{}
\newcommand{\joey}[1]{}
\newcommand{\dan}[1]{}
\newcommand{\kurt}[1]{}
\newcommand{\kevin}[1]{}
\newcommand{\sheng}[1]{}

\icmltitlerunning{Rethinking Model Size for Efficient Training and Inference of Transformers}
\begin{document}
\twocolumn[
\icmltitle{\textit{Train Large, Then Compress:} \\ Rethinking Model Size for Efficient Training and Inference of Transformers}

\icmlsetsymbol{equal}{*}

\begin{icmlauthorlist}
\icmlauthor{Zhuohan Li}{equal,berkeley}
\icmlauthor{Eric Wallace}{equal,berkeley}
\icmlauthor{Sheng Shen}{equal,berkeley}
\icmlauthor{Kevin Lin}{equal,berkeley}\\
\icmlauthor{Kurt Keutzer}{berkeley}
\icmlauthor{Dan Klein}{berkeley}
\icmlauthor{Joseph E. Gonzalez}{berkeley}
\end{icmlauthorlist}

\icmlaffiliation{berkeley}{UC Berkeley}

\icmlcorrespondingauthor{Zhuohan Li}{zhuohan@cs.berkeley.edu}

\icmlkeywords{NLP, BERT, Transformers, Efficient Training}
\vskip 0.3in
]

\printAffiliationsAndNotice{\icmlEqualContribution}
\begin{abstract}    
Since hardware resources are limited, the objective of training deep learning models is typically to maximize accuracy subject to the time and memory constraints of training and inference. We study the impact of model size in this setting, focusing on Transformer models for NLP tasks that are limited by compute: self-supervised pretraining and high-resource machine translation. We first show that even though smaller Transformer models execute faster per iteration, wider and deeper models converge in significantly fewer steps. Moreover, this acceleration in convergence typically outpaces the additional computational overhead of using larger models. Therefore, the most compute-efficient training strategy is to counterintuitively train extremely large models but stop after a small number of iterations.

This leads to an apparent trade-off between the training efficiency of large Transformer models and the inference efficiency of small Transformer models. However, we show that large models are more robust to compression techniques such as quantization and pruning than small models. Consequently, one can get the best of both worlds: heavily compressed, large models achieve higher accuracy than lightly compressed, small models.
\end{abstract}

\section{Introduction}
\begin{figure}[t]
\centering
\includegraphics[width=\linewidth, trim=2.1cm 9.7cm 2.1cm 0.2cm, clip]{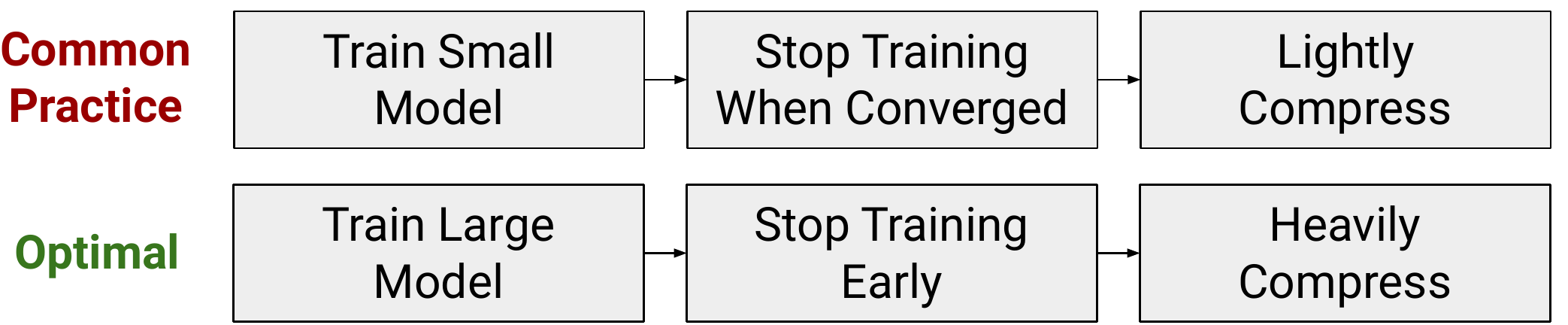}
\vspace{-0.43cm}
\caption{Under the usual presumption that models are trained to convergence, only small models that are fast-to-execute are feasible in resource-constrained settings. Our work shows that the most compute-efficient training scheme is instead to train very large models, stop them well short of convergence, and then heavily compress them to meet test-time constraints.}
\label{fig:flow_chart}
\end{figure}

\begin{figure*}[htp]
\centering
\subfigure[]{
    \includegraphics[width=0.43\textwidth]{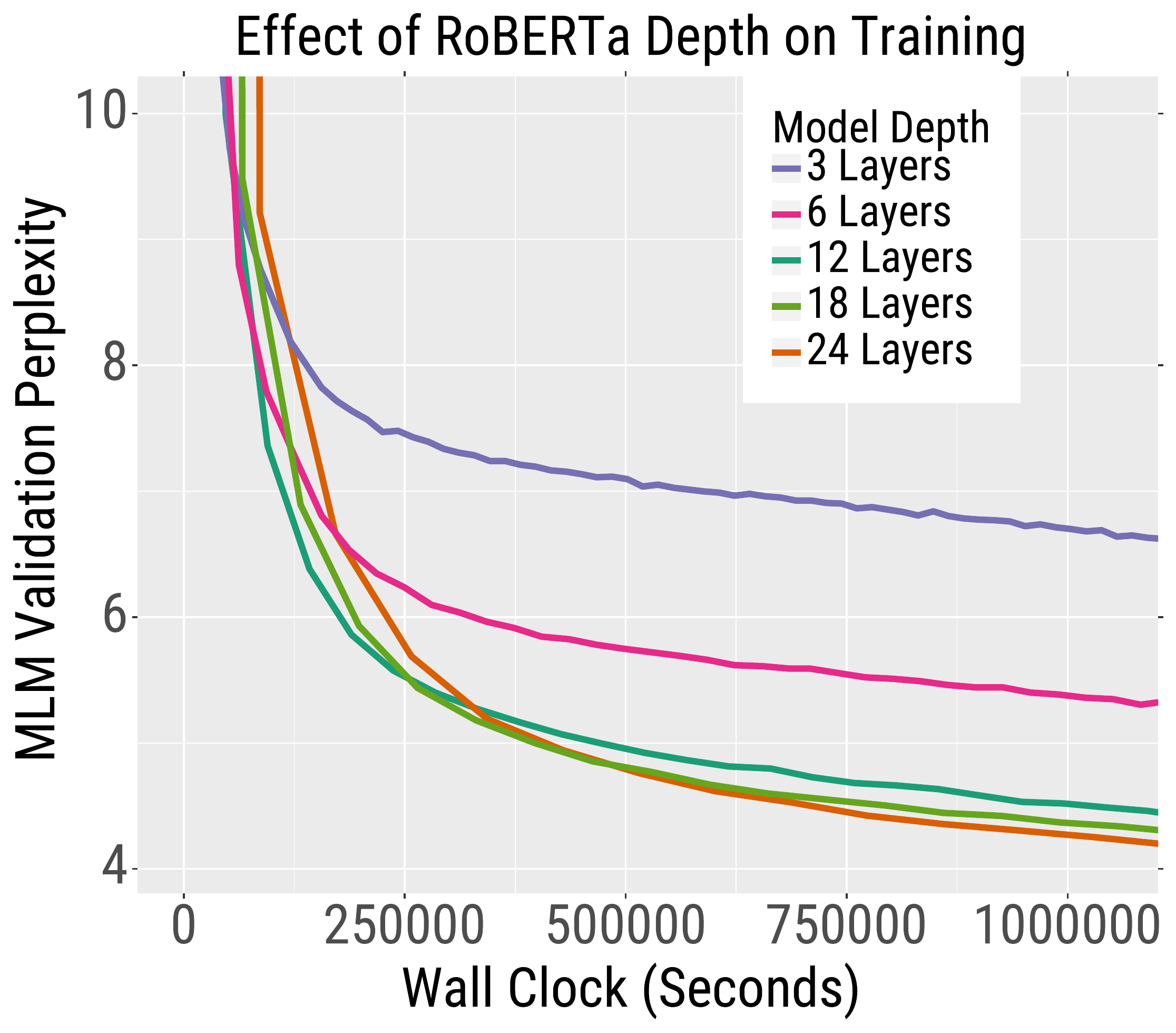}
    \label{fig:perplexity_vs_time}
    }
\subfigure[]{
\includegraphics[width=0.45\textwidth]{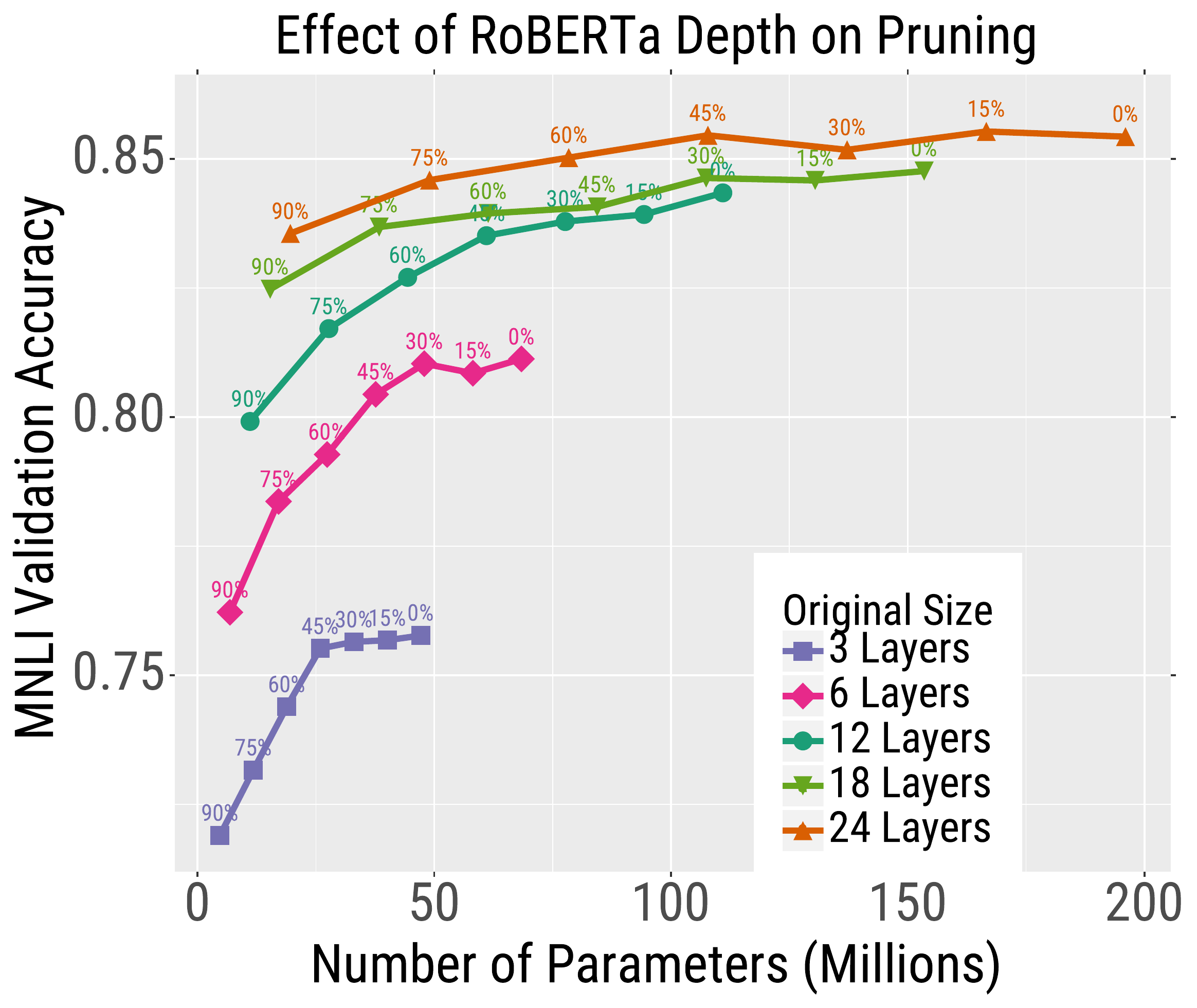}
  \label{fig:accuracy_vs_size}
} 
\vspace{-0.3cm}
\caption{Increasing Transformer model size results in lower validation error 
as a function of \textit{wall-clock time} and better test-time accuracy for a given \emph{inference budget}.
\textbf{\subref{fig:perplexity_vs_time}} demonstrates the training speedup for \roberta{} models of different sizes on the masked language modeling pretraining task. In  \textbf{\subref{fig:accuracy_vs_size}}, we take \roberta{} checkpoints that have been pretrained for the \textit{same} amount of wall-clock time and finetune them on a downstream dataset (MNLI). We then iteratively prune model weights to zero and find that the best models for a given test-time memory budget are ones which are trained large and then heavily compressed.}
\label{fig:intro}
\end{figure*}

In the current deep learning paradigm, using more compute (e.g., increasing model size, dataset size, or training steps) typically leads to higher model accuracy~\cite{brock2018large,raffel2019exploring}. This phenomenon is exacerbated by the recent success of self-supervised pretraining~\cite{devlin2018bert,hnaff2019dataefficient}, which allows training to scale to massive amounts of unlabeled data and very large neural models. Consequently, computational resources are increasingly the critical constraint on improving model accuracy. This constraint causes the (often implicit) goal of model training to be maximizing \textit{compute efficiency}: how to achieve the highest model accuracy given a fixed amount of hardware and training time.

Maximizing compute efficiency requires rethinking common assumptions about model training. In particular, there is typically an implicit assumption that models must be trained \textit{until convergence}, which makes larger models appear less viable for limited compute budgets. We challenge this assumption by demonstrating the opportunity to increase model size at the cost of convergence. Concretely, we show that the fastest way to train Transformer models~\cite{vaswani2017attention} is to substantially \emph{increase} model size but stop training very early. 

In our experiments, we vary the width and depth of Transformer models and evaluate their training time and accuracy on self-supervised pretraining (\roberta{}~\cite{liu2019roberta} trained on Wikipedia and BookCorpus) and machine translation (WMT14 English$\to$French). For these tasks, we first show that larger models converge to lower validation error in fewer gradient updates than smaller models (Section~\ref{sec:results}). Moreover, this increase in convergence outpaces the additional computational overhead of using larger models---the most compute-efficient models are extremely large and stopped well short of convergence (e.g., Figure~\ref{fig:intro}, left). We also show that this acceleration in wall-clock convergence is largely a function of parameter count and only weakly influenced by model width, depth, and batch size.

Although larger models train faster, they also increase the computational and memory requirements of inference.
This increased cost is especially problematic in real-world applications, where the cost of inference dominates the cost of training~\cite{jouppi2017datacenter,crankshaw2017clipper,metz2017tpu}. However, we show that for \roberta{}, this apparent trade-off can be reconciled with compression: large models are considerably more robust to compression as compared to small models (Section~\ref{sec:compression}). 
Thus, large, heavily compressed models outperform small, lightly compressed models using comparable inference costs (e.g., Figure~\ref{fig:intro}, right).

We finally analyze \textit{when} and \textit{why} large models train fast and compress well (Section~\ref{sec:why}). We show that the optimal model size is closely linked to the dataset size. In particular, large models perform favorably in big data settings where overfitting is a limited concern. We then analyze why larger models are more compressible by measuring the difference in weights when using quantized or sparse weight matrices. This error decreases as model size increases, i.e., greater overparameterization leads to easy-to-compress weights.
\section{Experimental Setup}

\subsection{Tasks, Models, and Datasets}

We train state-of-the-art models for two NLP tasks: self-supervised pretraining using masked language modeling and high-resource machine translation. We chose these tasks because accuracy continues to improve as models are made larger~\cite{shazeer2018mesh}, trained for more steps~\cite{liu2019roberta}, and trained using larger batches~\cite{raffel2019exploring}. Thus, a critical factor in improving accuracy for these tasks is to maximize the compute efficiency of training.

\paragraph{Self-supervised Pretraining (MLM)} We closely follow the pretraining setup and model from \roberta~\cite{liu2019roberta} with a few minor exceptions. We move the model's layer normalization layers~\cite{ba2016layer} to the input of every sub-layer (often called \emph{pre-norm}). This slightly improves results and stabilizes training~\cite{wang2019learning}. We also use an input sequence length of 128 and a batch size of 8192, unless otherwise noted. For \roberta{}, we vary the depth in $\{3, 6, 12, 18, 24\}$, and the hidden size in $\{256, 512, 768, 1024, 1536\}$.

The dataset for pretraining \roberta{} is not publicly available. We instead follow BERT~\cite{devlin2018bert} and concatenate the BookCorpus~\cite{zhu2015aligning} and a Wikipedia dump to use for training. Since the BookCorpus is no longer publicly available, we follow \citet{devlin2018bert} and crawl \url{http://smashwords.com}. Our final dataset is roughly 3.4 billion words in total. We hold out a random 0.5\% of the data for validation and report the masked language modeling (MLM) perplexity on this data. We also evaluate the model by finetuning on MNLI~\cite{williams2017broad} and SST-2~\cite{socher2013recursive}. We found the variance in accuracy for these two tasks to be lower than the other GLUE tasks~\cite{wang2018glue}.

\paragraph{Machine Translation} For machine translation (MT) we train the standard Transformer architecture and hyperparameters on the WMT14 English$\to$French dataset. We use the standard dataset splits: 36M sentences for training, \texttt{newstest2013} for validation, and \texttt{newstest2014} for testing. We follow standard practice and report tokenized case-sensitive BLEU~\cite{papineni2002bleu} with compound splitting~\cite{vaswani2017attention}. We vary the model depth in $\{2, 6, 8\}$ and hidden size in $\{128, 256, 512, 1024, 2048\}$.

\subsection{Evaluation Metrics: FLOPs and Wall-Clock Time}\label{subsec:hardware}
Recent work on resource-constrained training uses the total number of training steps~\cite{Li2020Budgeted} or the total number of training FLOPs~\cite{schwartz2019green,clark2020electra} as the main evaluation metric. These metrics do not adequately capture the true training time. In particular, reporting gradient steps does not account for the cost of using bigger batches or models. Moreover, although reporting FLOPs is useful for comparison as it is hardware-agnostic, it neglects the fact that parallel operations are significantly cheaper than sequential operations on modern hardware. 

We instead directly report wall-clock time as our main evaluation metric.\footnote{We also report selected learning curves as a function of FLOPs in Appendix~\ref{appendix:flops}. These curves show that our conclusion that larger models are faster to train is not specific to our hardware setup.} Since the runtime varies across machines (the hardware setups are different, the jobs are not isolated, etc.), we use a single machine to benchmark the time per gradient step for each model size. In particular, we train models and wait for the time per gradient step to stabilize, and then we use the average time over 100 steps to calculate the training duration. We conduct the timing on one NVIDIA 16GB V100 GPU and use gradient accumulation to fit larger models and batches. In order to be fair to smaller models, we increase the batch size to the largest size that fits in memory. This means that smaller models use fewer gradient accumulation steps and thus take less time per gradient step (which we confirmed empirically). We use  Tensor2Tensor~\cite{vaswani2018tensor2tensor} for MT and fairseq~\cite{ott2019fairseq} for RoBERTa. We train using a mix of v3-8 TPUs and 8xV100 GPUs for both tasks.
\section{Larger Models Train Faster}\label{sec:results}
 
\begin{figure}[t]
\centering
\includegraphics[trim={0cm 0.3cm 0cm 0.0cm},clip, width=0.9\columnwidth]{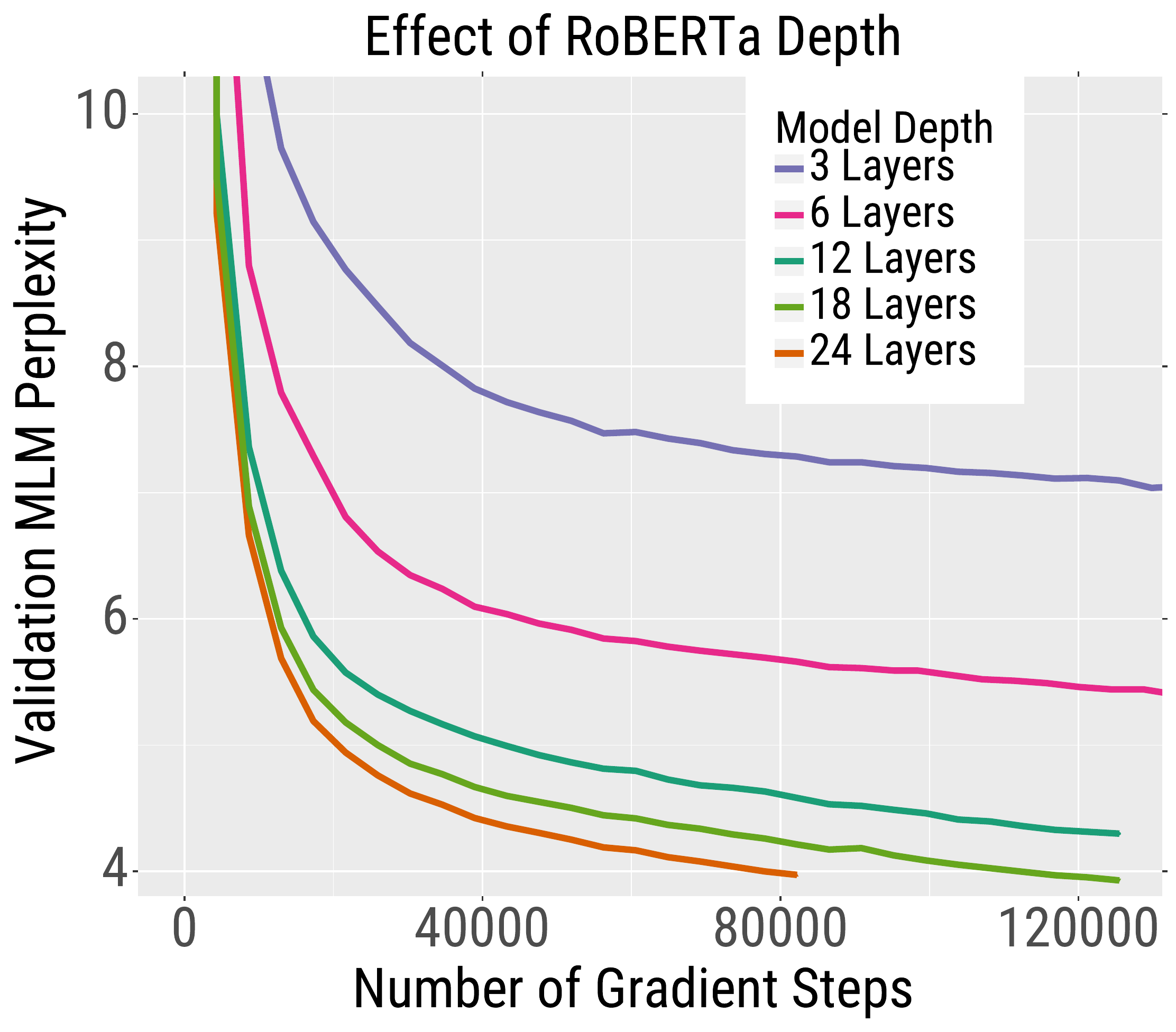}
\vspace{-0.3cm}
\caption{Deeper \roberta{} models converge faster than shallow models with respect to the gradient steps (wall-clock time shown in Figure~\ref{fig:intro}, left).}
\label{fig:depth}
\end{figure}

\begin{figure*}[t]
\centering
\hfill
\includegraphics[width=.39\textwidth]{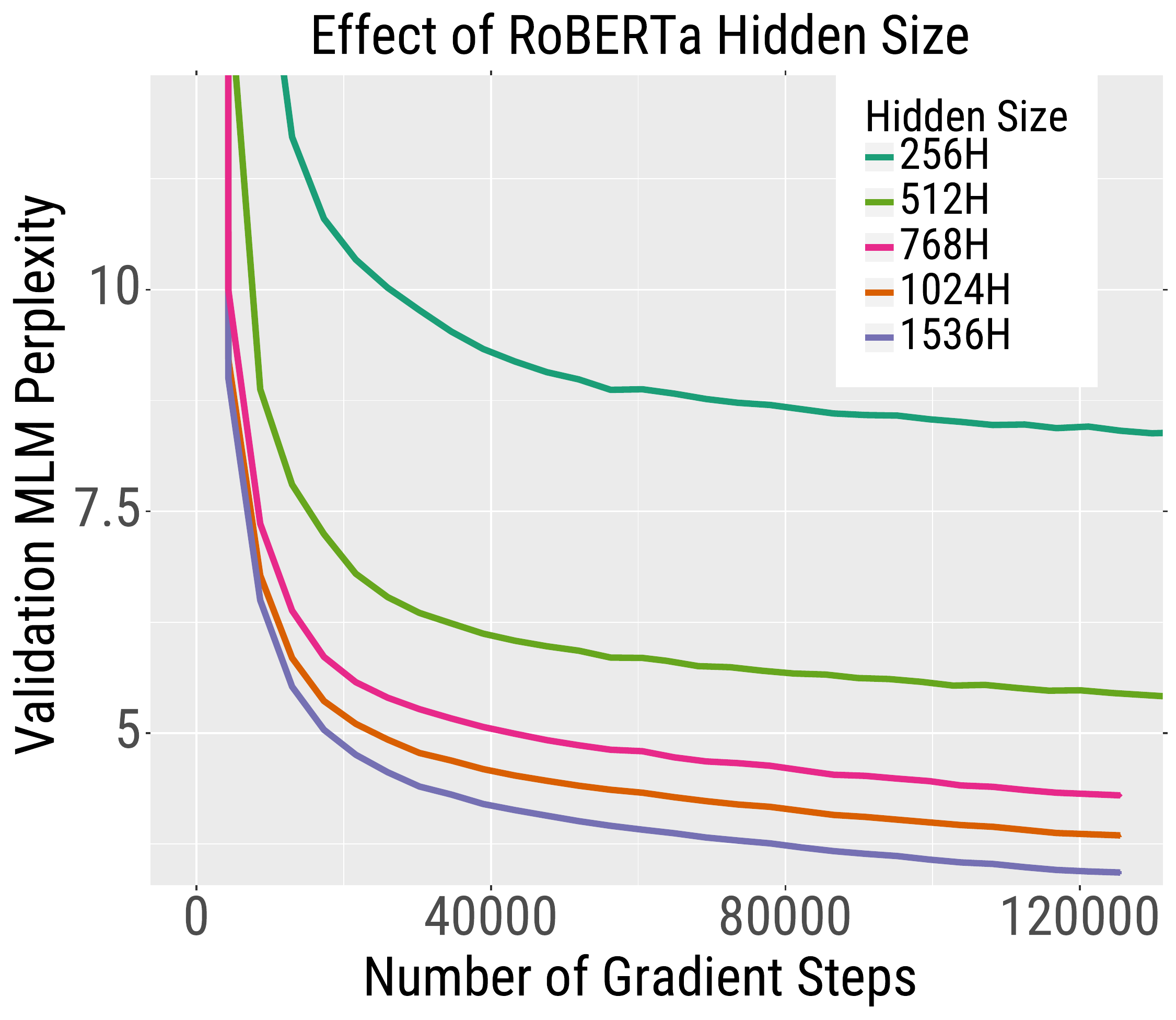}\hfill
\includegraphics[width=.406\textwidth]{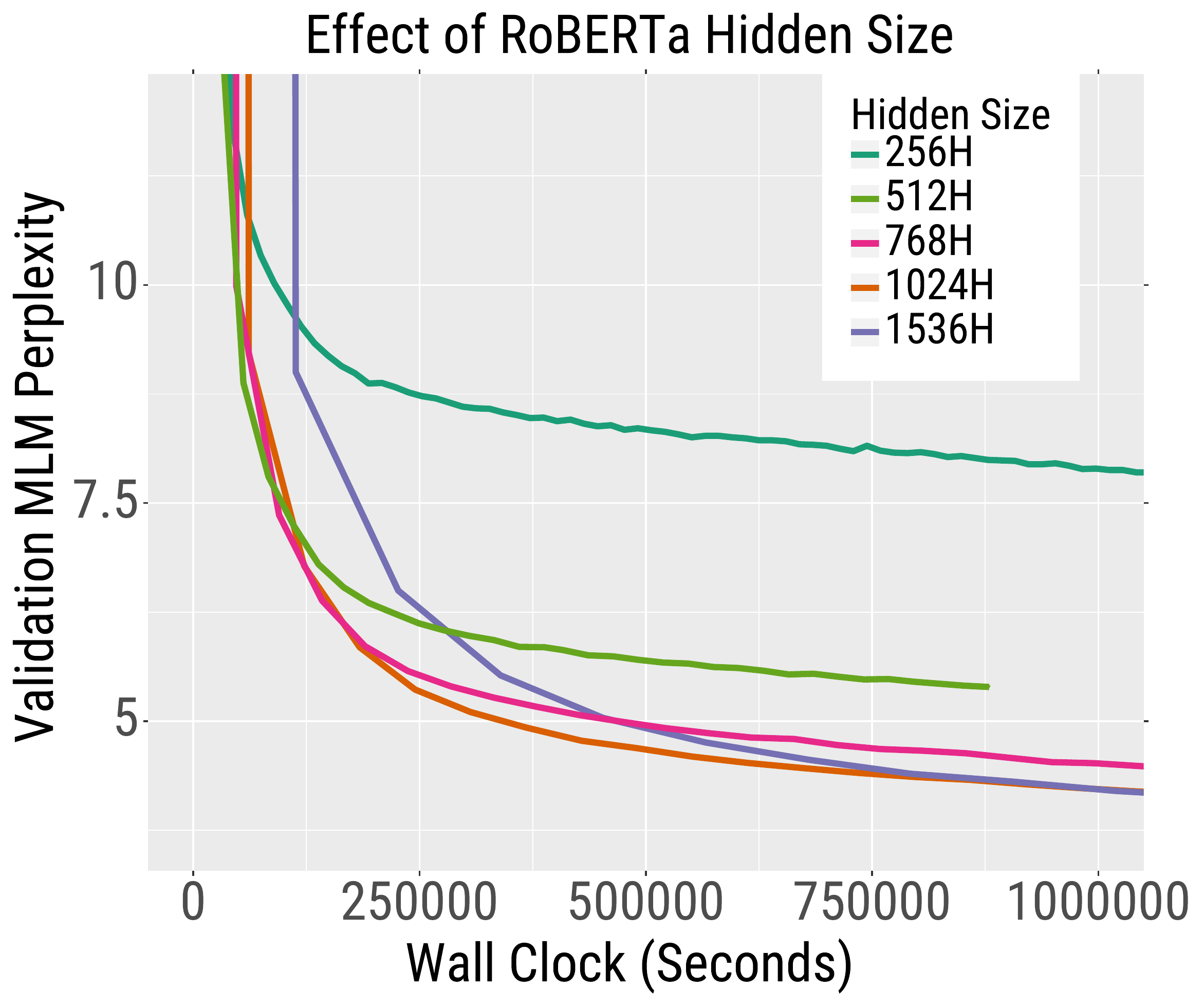}\hfill
\vspace{-0.3cm}
\caption{Wider models converge faster than narrower models as function of both gradient steps (left plot) and wall-clock time (right plot).}
\label{fig:width}
\end{figure*}

\begin{figure*}[t]
\centering
\hfill
\includegraphics[width=.41\textwidth]{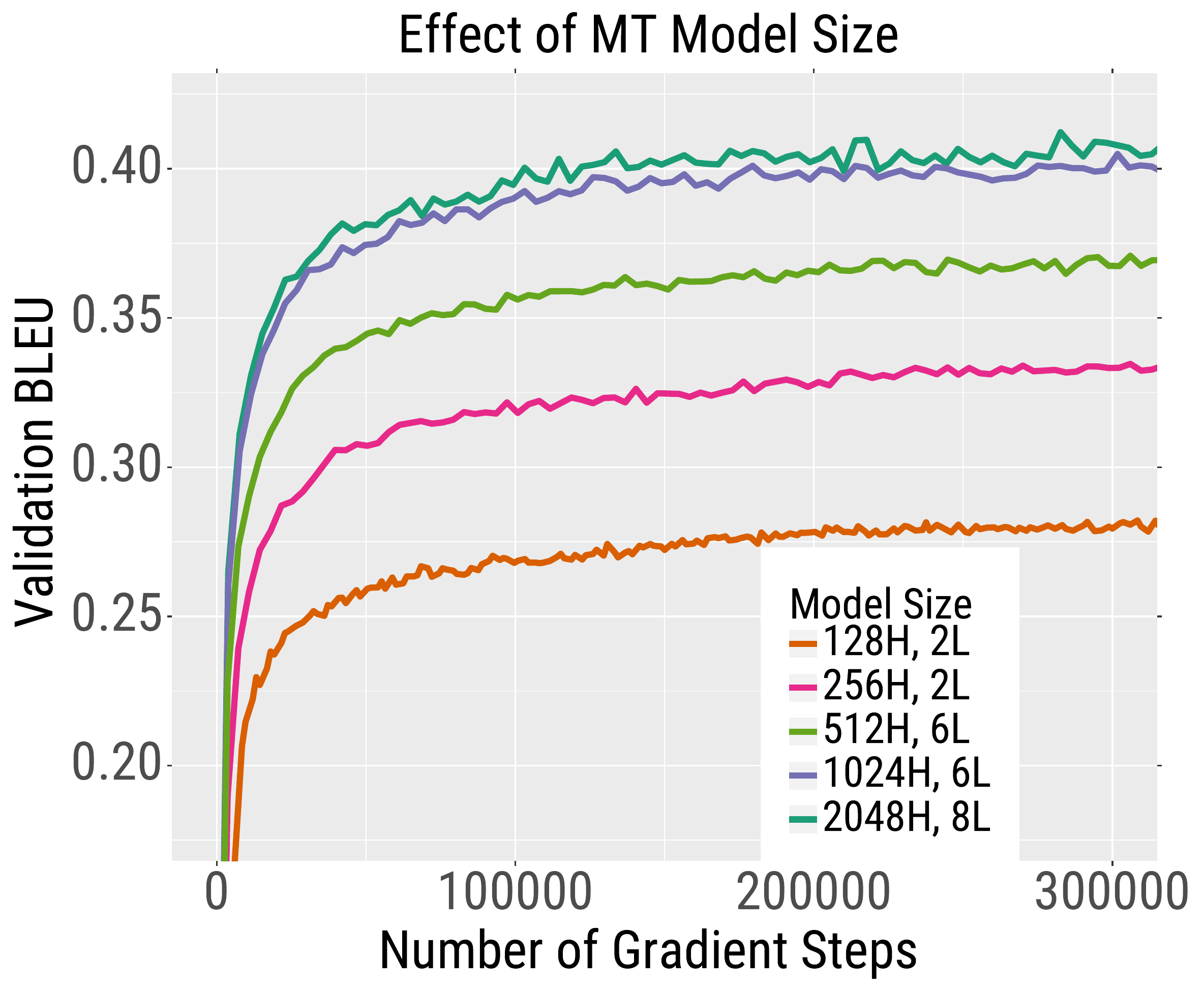}\hfill
\includegraphics[width=.4\textwidth]{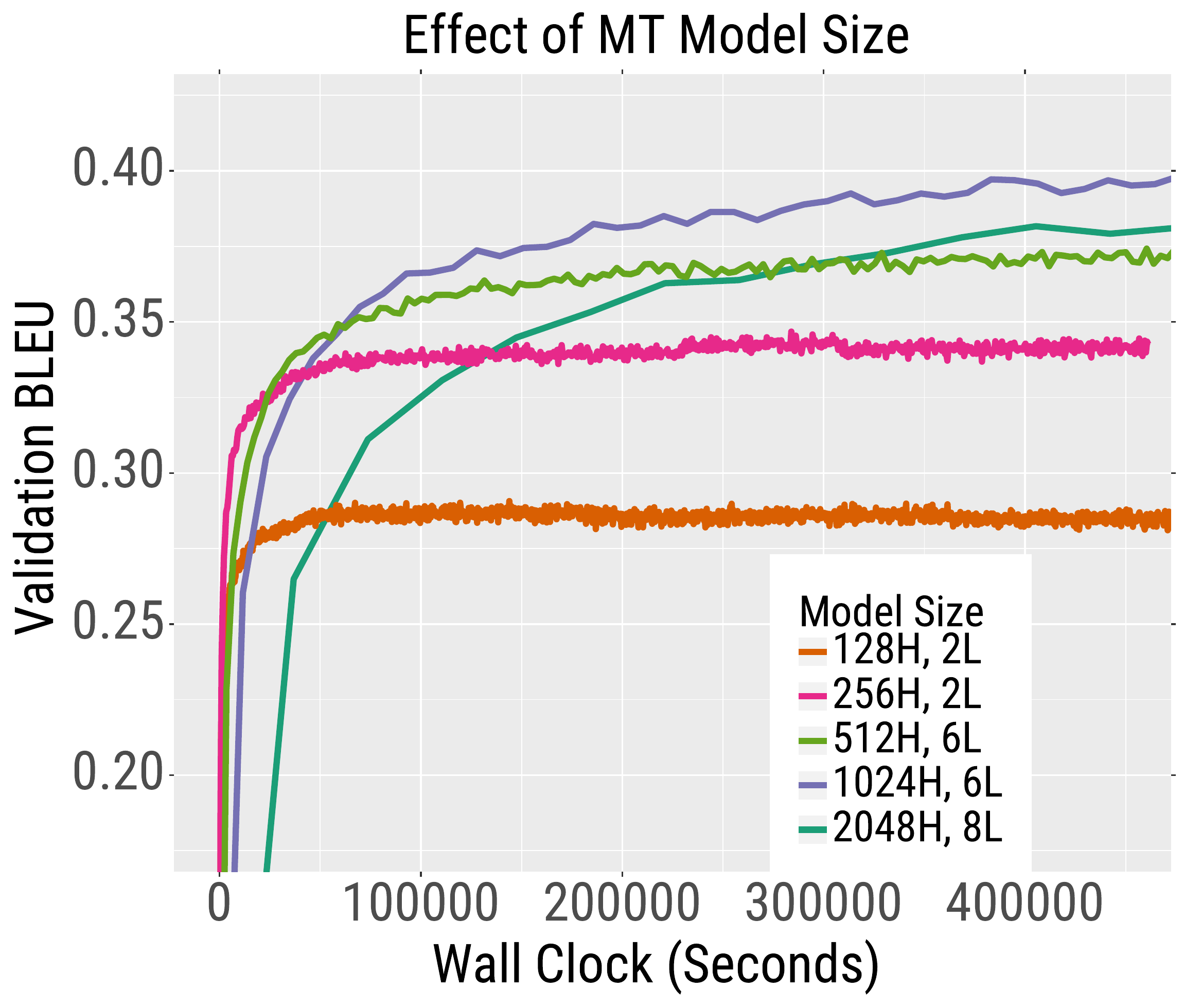}\hfill
\vspace{-0.3cm}
\caption{BLEU Scores on the English$\to$French validation set (\texttt{newstest2013}) using models of different sizes. Larger models typically converge faster as a function of both iterations (left plot) and wall-clock time (right plot). When models become too large (2048H, 6L), they converge faster per iteration but their overhead on our limited hardware negates their convergence improvements.}
\label{fig:mt}
\end{figure*}

Wider and deeper Transformer models are more sample-efficient than small models: they reach the same level of performance using fewer gradient steps (Figures~\ref{fig:depth}--\ref{fig:mt}). Moreover, this increase in convergence outpaces the additional computational overhead from increasing model size, even though we need to use more steps of gradient accumulation. Consequently, after adjusting for wall-clock time, the larger models are \emph{faster} to train than smaller models (Figures~\ref{fig:width}--\ref{fig:mt}).

\paragraph{Increase Model Width and Sometimes Depth} For the masked language modeling task, the validation perplexity weakly depends on the shape of the model. Instead, the total number of model parameters is the key determiner of the convergence rate. Thus, increasing either the width or the depth is effective at accelerating model training. On the other hand,  the preferred way to scale models for MT is to increase their width as wider models usually outperform deep models in final performance~\cite{vaswani2017attention,shazeer2018mesh}.\eric{just mention the parameter count and performance for a few models, because this is not shown by any figure at the moment.}

\paragraph{Increase Model Size, Not Batch Size} Another factor that affects the training efficiency is the batch size. In particular, there is a trade-off between using fast-to-execute small batches and slow-but-accurate large batches. We study the effect of scaling batch size because it provides an alternative to scaling model size. In particular, \textit{what if we use gradient accumulation to increase the batch size rather than the model size?} We vary the batch size for the 12 layer, 768H model and increase the learning rate as is common practice~\cite{goyal2017accurate,liu2019roberta}. We report the best found learning rate values in Table~\ref{table:learning_rates} in Appendix~\ref{appendix:curves}.

We show the training curves in Figure~\ref{fig:batch} in Appendix~\ref{appendix:curves}. Bigger batch sizes cause the model to converge in fewer steps. However, when adjusting for wall-clock time, increasing the batch size beyond a certain point only provides marginal improvements.\footnote{Note that our timing is done by accumulating gradients on a single GPU machine. For multi-GPU setups, the cost of accumulating gradients is lower as it naturally helps to balance out uneven runtimes across workers~\cite{ott2018scaling}. In this setup, the wall-clock improvements from increasing batch sizes by accumulating gradients may be slightly larger.} In particular, varying the batch size has little impact when training with a batch size in the range from 2048--16384. This aligns with the findings of \citet{mccandlish2018empirical}: training efficiency is maximized when models are trained near some \emph{critical batch size}. 

An additional downside of increasing the batch size is that it requires simultaneously tuning the learning rate. On the other hand, scaling model size provides improvements in training efficiency \textit{without} adjusting any hyperparameters. Overall, our results show that one should increase the batch size (and learning rate) until the critical batch size region is reached and then to focus on increasing model size.

\paragraph{Larger Models Are Not Harder to Finetune} Although the larger models minimize validation MLM perplexity faster, one concern is that they may not minimize downstream task error faster. For instance, larger models may overfit on small downstream datasets. We investigate this by training \roberta{} models of different sizes and stopping them when they reach the same MLM perplexity (the larger models have been trained for \textit{less} wall-clock time). We then finetune each model using the \roberta{} finetuning hyperparameters~\cite{liu2019roberta} on MNLI and SST-2. We report the model accuracies in Table~\ref{table:bigger_finetuning} in Appendix~\ref{appendix:bigger_finetuning}. All models reach comparable accuracies (in fact, the larger models typically outperform the smaller ones), which shows that larger models are not more difficult to finetune.

\paragraph{Returns Diminish As Size Increases} For both RoBERTa and MT, the largest models have reached the point where they stop improving convergence with respect to wall-clock time. For example, the largest model for MT (6L, 2048H) starts to converge slower with respect to wall-clock time than the second-largest model (6L, 1024H). These diminishing returns occur because (1) the per-step convergence improvements from using larger models decreases as the model gets larger and (2) the computational overhead increases as our hardware becomes increasingly compute-bound. We further analyze when and why returns diminish in Section~\ref{sec:why}.
\section{Larger Models Compress Better}\label{sec:compression}
Although the most compute-efficient \emph{training} scheme is to use larger models, this results in models which are less \emph{inference} efficient. Here, we demonstrate how to get the best of both worlds. In particular, we show that since large models are more compressible than small models, they can outperform small models while using similar inference costs.

\subsection{Compression Methodology and Evaluation}

\paragraph{Compression Methods}
Model compression methods reduce the inference costs of trained models. For example, model compression can reduce inference latency to enable real-time applications like simultaneous MT~\cite{see2016compression} or reduce memory usage to save energy for mobile devices~\cite{han2015deep}. We focus on compression methods which are \emph{fast} to perform---methods which require significant amounts of compute will negate the speedup from using larger models.\footnote{For example, we avoid using model distillation methods because they can add a significant computational overhead~\cite{sanh2019distilbert,turc2019well} or cause a significant degradation in accuracy~\cite{liu2019attentive,sun2019patient}.} In particular, we consider two compression techniques: quantization (Section~\ref{subsec:quantization}) and pruning (Section~\ref{subsec:pruning}), as well as their combination.\footnote{We also experiment with \emph{parameter sharing}~\cite{lan2019albert,dehghani2018universal}---tying the weights of the Transformer layers together---and find that it slows convergence (see Appendix~\ref{appendix:sharing}).} Quantization stores model weights in low precision formats to (1) accelerate operations when using hardware with reduced precision support and (2) reduce overall memory footprint~\cite{han2015deep,dong2019hawq}. Pruning sets neural network weights to zero to (1) remove operations and (2) reduce the memory footprint when models are stored in sparse matrix formats~\cite{lecun1990optimal,han2015learning}. We apply both quantization and pruning post-hoc to the finetuned models to limit the additional computational overhead.

\paragraph{Finetuning Setup and Compression Evaluation} We focus on compressing the finetuned \roberta{} models as a case study. We train models of different sizes for 1,000,000 seconds,\footnote{We expect similar conclusions to hold for other budgets.} finetune them on MNLI/SST-2, and then apply quantization/pruning. For evaluation, even though pruning and quantization will improve inference latency/throughput, quantifying these improvements is challenging because they are highly hardware-dependent. Instead, we follow past work and report the memory needed to store the model parameters~\cite{thakker2019compressing,shen2019q}.

\subsection{Larger Models Are More Robust to Quantization}\label{subsec:quantization}

\begin{figure*}[htp]
\centering
\hfill
\includegraphics[width=.41\textwidth]{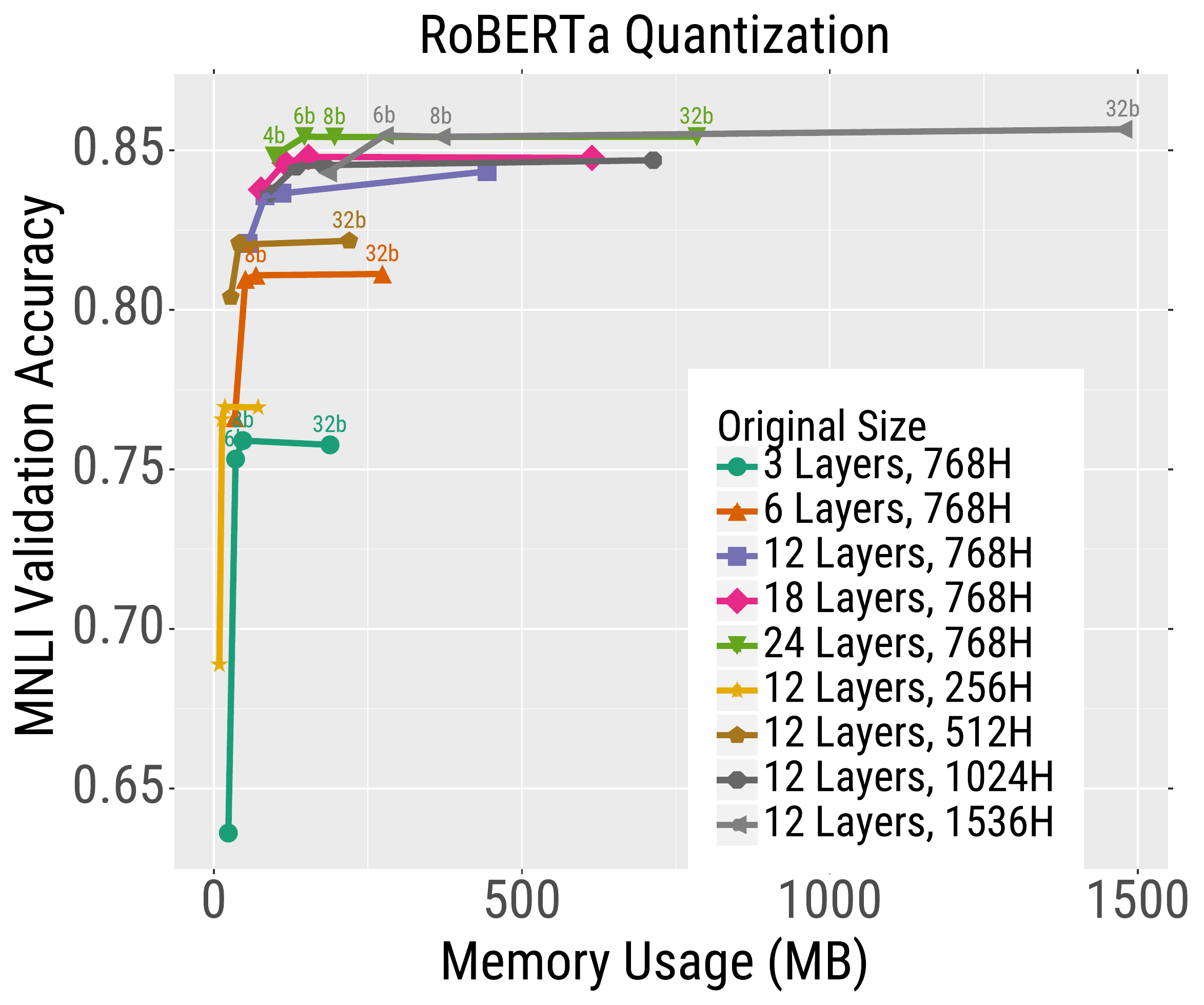}\hfill
\includegraphics[width=.4\textwidth]{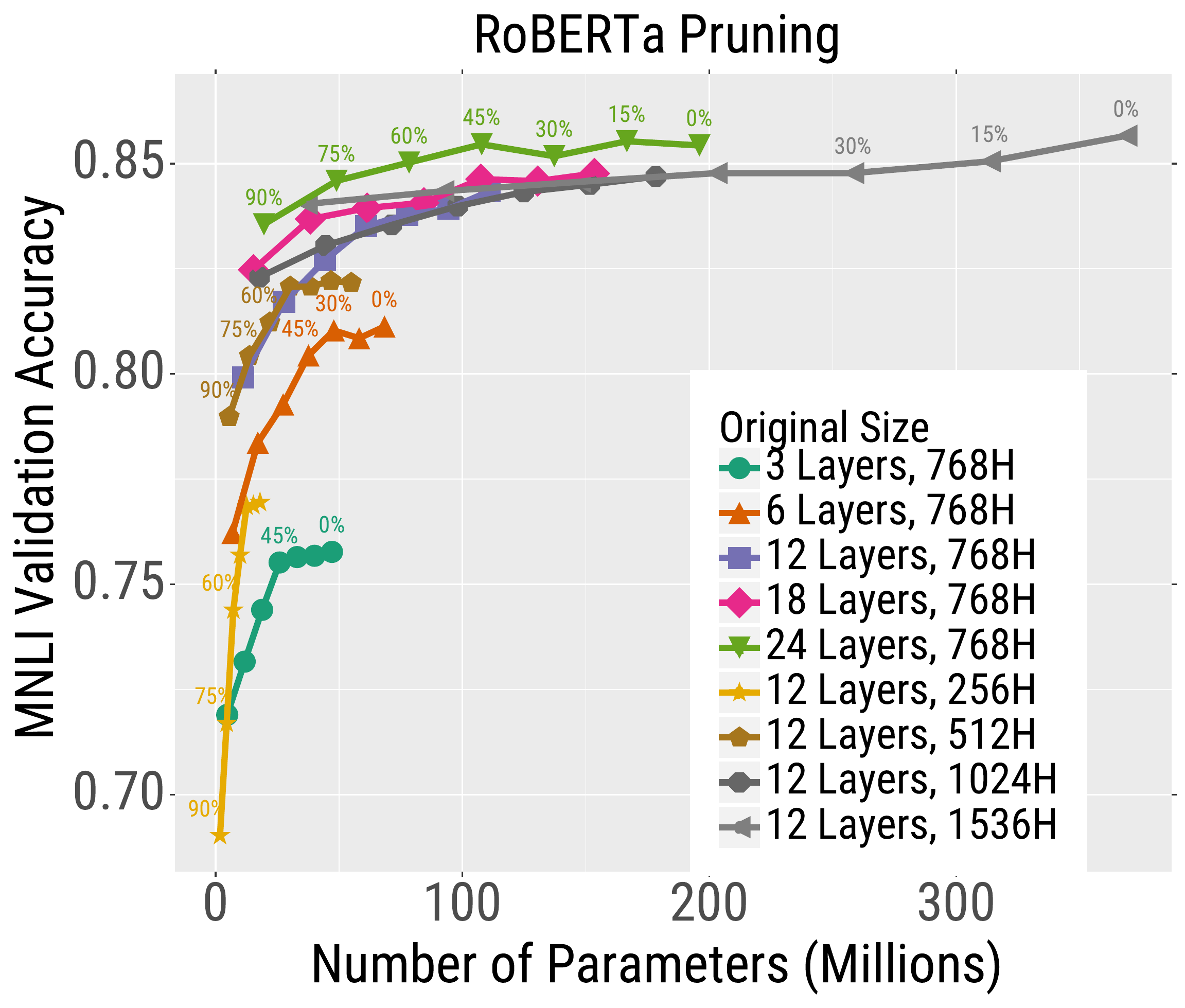}\hfill
\vspace{-0.3cm}
\caption{We first pretrain \roberta{} models of different sizes for the \emph{same} total wall-clock time (larger models are trained for fewer steps). We then finetune each model on MNLI and compress them using quantization (left) and pruning (right). For most budgets (x-axis), the highest accuracy models are the ones which are trained large and then heavily compressed. The labels above each point indicate the compression amount (e.g., 4-bit quantization or 45\% sparsity); we omit cluttered labels. SST-2 results are shown in Appendix~\ref{appendix:compression}.}
\label{fig:compression}
\end{figure*}

We quantize every parameter, including the embedding matrix, but keep the model activations at full precision. We use floating point precisions in $\{4, 6, 8, 32\}$ bits (using lower than 4-bits resulted in severe accuracy loss). We apply quantization post-hoc which adds \textit{no} additional time.

We quantize uniformly: the range of floats is equally split and represented by unsigned integers in $\{0, \ldots, 2^k - 1\}$, where $k$ is the precision. We accomplish this by quantizing the weights $W$ as:
\setlength{\abovedisplayskip}{7pt}
\setlength{\belowdisplayskip}{7pt}
\begin{align*}
&W^\prime = \texttt{Clamp}(W, q_0, q_{2^{k}-1}), \\
&W^I = \lfloor \frac{W^\prime - q_0}{\Delta} \rceil, \text{ where } \Delta = \frac{q_{2^{k}-1} - q_0}{2^k - 1}, \\
&\texttt{Quantize}(W) = \Delta W^I + q_0,
\end{align*}
where \texttt{Clamp()} clamps all elements to the min/max range, $W^I$ is a set of integer indices, $\lfloor \cdot \rceil$ is the round operator, $\Delta$ is the distance between two adjacent quantized points, and $[q_0, q_{2^{k}- 1}]$ indicates the quantization range.

\paragraph{Results} The quantization results for MNLI are shown on the left of Figure~\ref{fig:compression} (SST-2 results are in Appendix~\ref{appendix:compression}). We plot each model's accuracy at different quantization levels as a function of its total memory usage. The larger models are more robust to quantization than the smaller models (the accuracy drop is smaller when the precision is reduced). Hence, the models which are trained using large parameter counts and then heavily quantized achieve the highest accuracy for almost all memory budgets.

\subsection{Larger Models Are More Robust to Pruning}\label{subsec:pruning}

We use iterative magnitude pruning~\cite{strom1997sparse,han2015deep}: we iteratively zero out the smallest magnitude parameters and continue finetuning the model on the downstream task to recover lost accuracy.

Concretely, we consider models with sparsity levels of 15\%, 30\%, 45\%, 60\%, 75\%, and 90\%. We first find the 15\% of weights with the smallest magnitude and set them to zero.\footnote{It also may be possible to remove entire attention heads in addition to zeroing out weights~\cite{michel2019sixteen,voita2019analyzing}. This may further improve our compression results.} We then finetune the model on the downstream task until it reaches within 99.5\% of its original validation accuracy or until we reach one training epoch. We then repeat this process---we prune another 15\% of the smallest magnitude weights and finetune---stopping when we reach the desired sparsity level. The additional training overhead from this iterative process is small because the model typically recovers its accuracy in significantly less than one epoch (sometimes it does not require any retraining to maintain 99.5\%). For example, pruning to 45\% can be done with one or two additional epochs of finetuning on MNLI.

\paragraph{Results} The pruning results for MNLI are shown in the right of Figure~\ref{fig:compression}. We report the model's accuracy as a function of the total number of nonzero parameters.\footnote{Since the reduction in memory from storing sparse matrices is highly dependent on the data structure used, we follow past work and report the number of nonzero model parameters~\cite{luo2017thinet, li2016pruning}.} The larger models can be pruned more than the smaller models without significantly hurting accuracy. Consequently, the large, heavily pruned models provide the best accuracy-efficiency trade-off. We find that deep networks are more robust to pruning than wider networks, e.g., the 24 Layer, 768H model outperforms the 12 Layer, 1536H model at most test budgets.

\begin{figure}[t]
\centering
\includegraphics[trim={0cm 0.3cm 0cm 0.0cm},clip, width=0.9\columnwidth]{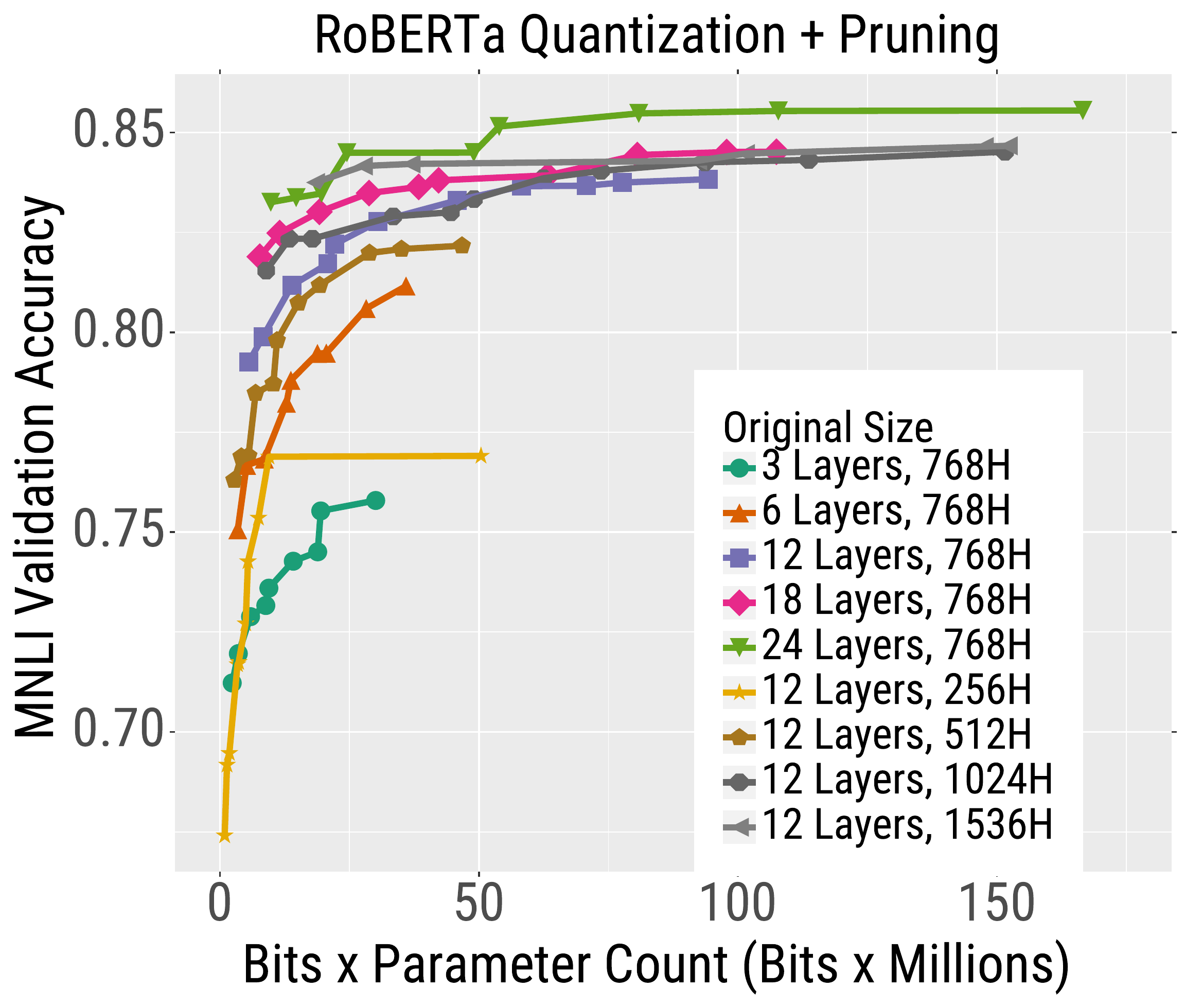}
\vspace{-0.2cm}
\caption{We combine pruning and quantization and find their gains to be complementary. The models which are trained large and then compressed are the best performing for each test-time budget.}
\label{fig:combining}
\end{figure}

\paragraph{Combining Quantization and Pruning Results} Pruning and quantization are complementary techniques for compressing Transformer models. We first prune models to various sparsity levels (e.g., 15\%, 30\%, etc.) and then apply varying amounts of quantization (e.g., 8-bit, 4-bit, etc.) to each model. In Figure~\ref{fig:combining} we plot combinations of pruning and quantization that lie at or near the Pareto frontier. Large models that are heavily compressed still provide the best trade-off between accuracy and efficiency when leveraging both pruning and quantization. A particularly strong compression method is to prune 30-40\% of the weights and then quantize the model to 6-8 bits.

\subsection{Convergence Does Not Affect Compressibility}

Although larger Transformer models are more compressible, there is a confounding factor that our larger models are also less \textit{converged} on the pretraining task. Is it the larger model size or the lack of convergence that causes the enhanced compressibility? We investigate this by finetuning \roberta{} models starting from different pretraining checkpoints (e.g., 3 epochs, 6 epochs, etc.) on MNLI. We then quantize the models to 4-bits.

\begin{figure}[t]
\centering
\includegraphics[trim={0cm 0.3cm 0cm 0.0cm},clip, width=0.9\columnwidth]{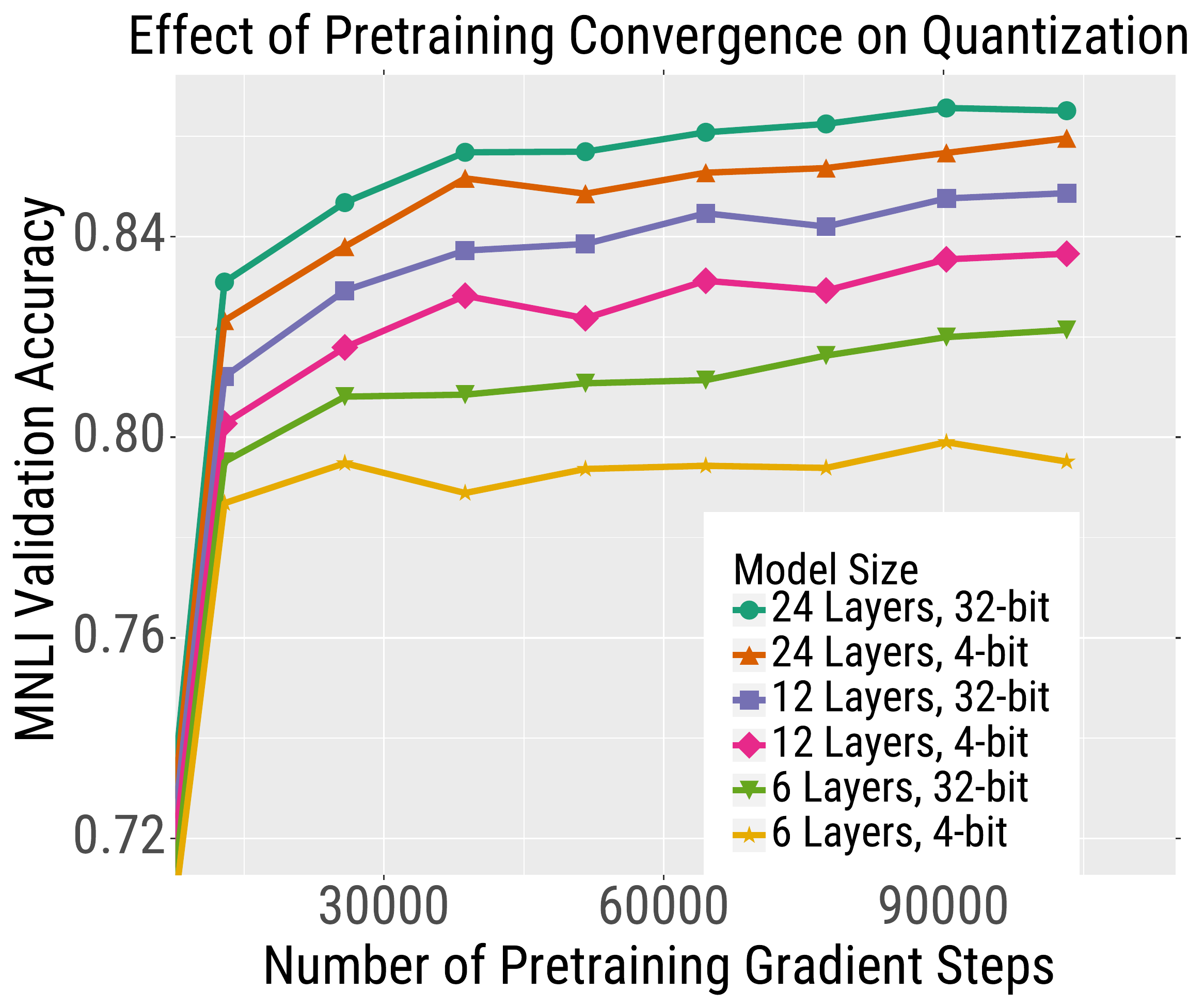}
\vspace{-0.2cm}
\caption{We disentangle whether \textit{model size} or \textit{pretraining convergence} causes the enhanced compressibility of larger models. We finetune \roberta{} models starting from different pretraining checkpoints on MNLI. We then quantize the models to 4-bits. Quantization is hardly affected by convergence---the drop in MNLI accuracy due to quantization is comparable as the pretrained model becomes more converged. Instead, the factor that determines compressibility is model size---the drop in accuracy is very large when compressing smaller models and vice versa.}
\label{fig:convergence}
\end{figure}

Figure~\ref{fig:convergence} shows the results. Quantization is hardly affected by pretraining convergence---the drop in accuracy between the full precision and the 4-bit precision MNLI models is comparable as the pretrained model becomes more converged. Instead, the factor that determines compressibility is model size---the drop in accuracy is very large when compressing smaller models and vice versa.

\section{When and Why Are Larger Models Better?}\label{sec:why}

This section presents results and discussion on why larger Transformer models train faster and compress better.

\subsection{Better Sample Efficiency With Larger Models}

For larger models to train faster, they must converge faster (w.r.t. test error) per iteration. While there is a robust literature studying why larger models achieve better \emph{final test accuracy},\footnote{Chiefly, this work seeks to reconcile the conflict between modern deep learning practice and the classical bias-variance trade-off. For instance, it studies forms of implicit regularization~\cite{zhang2016understanding,belkin2018reconciling}, characterizes the expressivity of deep models~\cite{raghu2017expressive,lu2017expressive}, and bounds the neural network generalization error~\cite{du2018gradient,arora2018stronger}.} there is considerably less work exploring if and why larger models converge faster. One initial step in this direction is \citet{arora2018optimization}, who show that for deep \textit{linear} neural networks, increasing depth can promote movement along directions already taken by the optimizer. 

\paragraph{Fast Minimization and the Role of Overfitting} 
One empirical reason for the acceleration in convergence is that larger Transformer models minimize the training error faster. And, since the generalization gap is small for our tasks due to very large training sets, the larger models also converge faster w.r.t test error. In fact, the challenge in the MLM task is not \emph{overfitting}, but instead, it is \emph{fitting} the data---even 8 billion parameter models do not overfit to large pretraining corpora~\cite{shoeybi2019megatron}. 

When overfitting \textit{is} a concern, larger models start to converge slower (w.r.t test error). We demonstrate this by randomly subsampling our pretraining dataset to 5\% and 1\% of its original size and training \roberta{} models of various sizes. When subsampling the data to 5\% (top row of Figure~\ref{fig:subsample_data} in Appendix~\ref{appendix:curves}), the largest models do not improve on the training time of the smaller models (e.g., 12 layer \roberta{} trains just as fast as a 24 layer \roberta{}). Moreover, when the data is subsampled to 1\% (bottom row of Figure~\ref{fig:subsample_data}), the largest models are worse in terms of perplexity due to overfitting. Thus, although our main conclusion that increasing model size accelerates convergence still holds for the smaller models (e.g., the 12 layer model outperforms the 3 layer one), overfitting causes it to break down for the largest models.

\subsection{Manageable Compute Costs for Large Models}

For larger models to train faster with respect to wall-clock time, their convergence improvements must not be negated by their slowdown in per-iteration time. Fortunately, parallel hardware (e.g., GPUs, TPUs) is usually not compute bound when training deep learning models. Instead, memory storage/movement is the limiting factor in image classification~\cite{gomez2017reversible}, semantic segmentation~\cite{chen2017deeplab}, language modeling~\cite{kitaev2020reformer}, and other tasks~\cite{jain2019checkmate}. Thus, larger models will more fully utilize the available compute, causing their slowdown to be sublinear. Moreover, when larger models cause hardware to run out of memory, gradient accumulation can trade-off memory for compute while still preserving the gains of large models, as shown in our experiments.

\subsection{Smaller Compression Error for Larger Models}

\begin{figure*}[htp]
\centering
\hfill
\includegraphics[width=.485\textwidth]{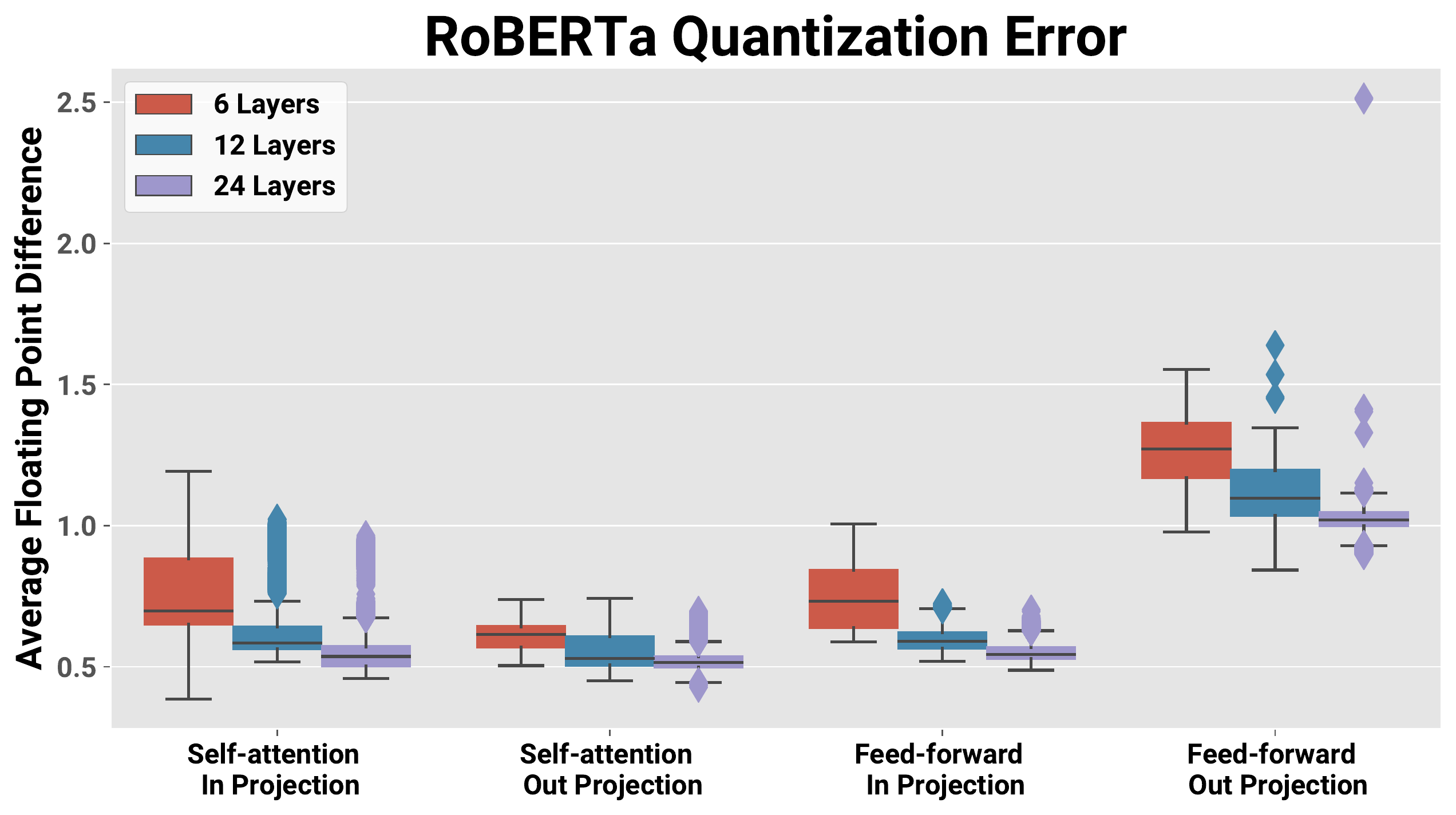}\hfill
\includegraphics[width=.485\textwidth]{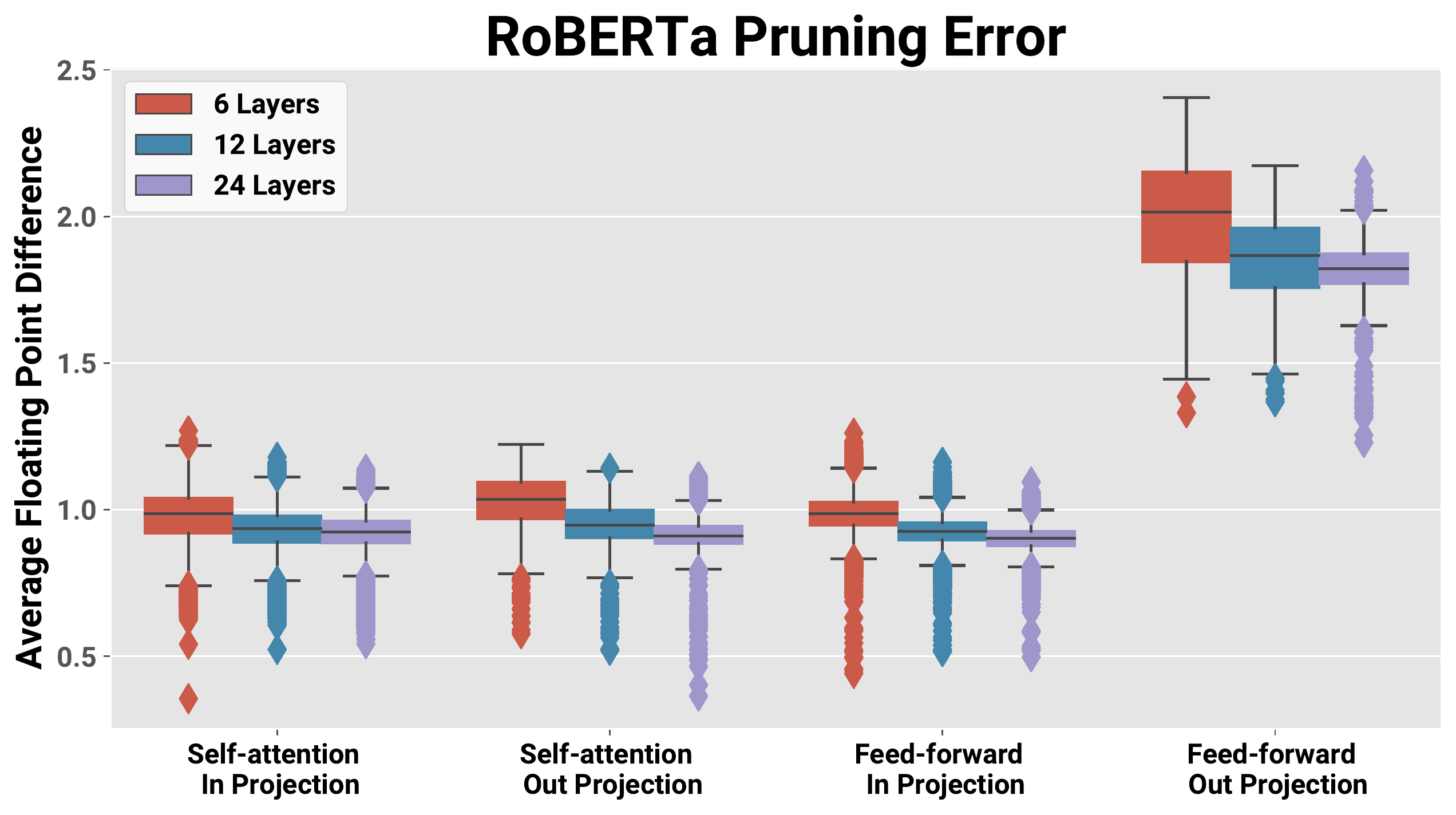}\hfill
\vspace{-0.3cm}
\caption{We finetune \roberta{} models of different sizes (6 layers, 12 layers, and 24 layers) on MNLI. We then quantize models to 4-bits or prune models to 60\% sparsity. We plot the difference between the weights of the original and the quantized/pruned models averaged across different modules in the Transformer. The mean and variance of the weight difference after quantization (left) is consistently lower for the deeper models compared to the shallower models. The same holds for the difference after pruning (right). This shows that the larger model's weights are naturally easier to approximate with low-precision / sparse matrices than smaller models.}
\label{fig:quantization_error}
\end{figure*}

Large transformer models are more compressible than small transformer models.\footnote{Similar findings hold for large but sparse audio synthesis models~\cite{kalchbrenner2018efficient} and convolutional models for computer vision~\cite{zhu2017prune,elsen2019fast,evci2019rigging,kusupati2020soft}.} Here, we present initial experiments to better understand why this occurs.
 
\paragraph{Quantization Error is Smaller for Larger Models}  We first measure the \textit{quantization error}---the difference between the full-precision and low-precision weights---for the 4-bit \roberta{} models. On the left of Figure~\ref{fig:quantization_error}, we plot this value for models of varying depths (6, 12, and 24 layers) averaged across different Transformer modules (e.g., in-projection matrix of the self-attention). The mean and variance of the quantization error are smaller for deeper models.

\paragraph{Pruning Error is Smaller for Larger Models} Similarly, we measure the \emph{pruning error}---the difference between the original weights and the sparse weights---for the 60\% sparse \roberta{} models. The mean and variance of the pruning error are smaller for deeper models (Figure~\ref{fig:quantization_error}, right).

These two results show that the larger model's weights are more easily approximated by low-precision or sparse matrices. Interestingly, this phenomenon naturally occurs without directly optimizing for it; an area for future work is to study why these weight patterns emerge in larger models. 

\paragraph{Connection to the Lottery Ticket Hypothesis} Our compression findings have deep connections to recent conjectures such as the lottery ticket hypothesis~\cite{frankle2018lottery}. The lottery ticket hypothesis argues that larger models are preferable as they have a higher chance of finding a lucky initialization in one of their subnetworks. Our work shows that, for certain accuracies, as models become \textit{increasingly large}, they contain \textit{increasingly small} subnetworks which achieve that accuracy.

\section{Related Work}\label{sec:related}

\paragraph{Improving Training Speed and Efficiency} There is a large body of work on accelerating model training, traditionally accomplished via improved optimizers~\cite{nesterov,adam}. More recent work improves training efficiency by modifying loss functions~\cite{clark2020electra}, model structures/sparsities~\cite{louizos2018learning,gong2019efficient,tan2019efficientnet}, backpropagation storage requirements~\cite{gruslys2016memory}, or learning rate schedules~\cite{loshchilov2016sgdr,Li2020Budgeted}. We study the impact of model size, which is largely orthogonal to these other training efficiency improvements.~\smallskip

\noindent \textbf{Scaling Model Training} Another line of work scales model training to large amounts of distributed hardware and addresses the associated systems and machine learning challenges~\cite{goyal2017accurate,ott2018scaling,you2019reducing}. Our work instead looks to choose the optimal model size for a fixed (small) hardware budget. Future work can study whether our conclusion that large models are more compute-efficient also holds in this highly-distributed setting, where the ``budget'' is extremely large.~\smallskip

\noindent \textbf{Hyperparameter Tuning and AutoML} In our work, we have an initial setting for the  hyperparameters and optimize the model size. However, good initial models and hyperparameters are unknown when approaching \textit{new} problems. For these cases, the optimal training strategy must consider the cost of experimenting with different architectures and hyperparameters; future work can study the effect of model size in this setting. More generally, our findings may impact the design of automated methods for solving/optimizing machine learning problems~\cite{feurer2015efficient,zoph2016neural,jaderberg2017population}. In particular, the compute-efficiency of these methods may improve by following our \emph{train large, then compress} methodology.~\smallskip

\noindent \textbf{Training Efficiency of Large Models} Recent and concurrent work also considers the impact of model size on the compute efficiency of training. \citet{raffel2019exploring} show that training a 4x larger Transformer model is a good usage of 4x more compute. \citet{ardalani2019empirically} show that larger RNN models take fewer gradient iterations to converge but do not consider that larger models are faster when adjusting for wall-clock time. In concurrent work, \citet{kaplan2020scaling} study the impact of model size on the training efficiency of Transformer language models. They make similar conclusions that large, undertrained models are superior to small, well-trained models. Our work differs in that we study machine translation and the impact of training large models on downstream tasks (model finetuning and compression).
\newpage
\section{Conclusion and Future Work}\label{sec:conclusion}

We studied the impact of Transformer model size on the efficiency of training and inference. We show that increasing model width and depth accelerates convergence in terms of both gradient steps and wall-clock time. Moreover, even though large models appear less efficient during inference, we demonstrate that they are more robust to compression. Therefore, we conclude that the best strategy for resource-constrained training is to \emph{train large models and then heavily compress them}. 

In the future, we will examine these conclusions on more domains such as computer vision. Moreover, we look to answer the questions that are raised by our results: \textit{why} do larger transformer models train fast and compress well, how does model size impact overfitting and hyperparameter tuning, and more generally, what other common design decisions should be rethought in the compute-efficient setting?
\section*{Acknowledgements}
This research was supported by the Berkeley RISE Lab. We would like to thank the Google Cloud TPU team for their hardware support. We are also grateful to Shi Feng, Yang Liu, Suchin Gururangan, Nelson Liu, the members of Berkeley NLP, and the members of the Berkeley RISE Lab for their valuable feedback.

\bibliography{journal-abbrv,bib}
\bibliographystyle{icml2020}

\appendix
\clearpage
\clearpage

\section{Additional Training Curves}\label{appendix:curves}

\subsection{Training Cost Using FLOPs}\label{appendix:flops}

\begin{figure*}[t]
\centering
\hfill
\includegraphics[width=.325\textwidth]{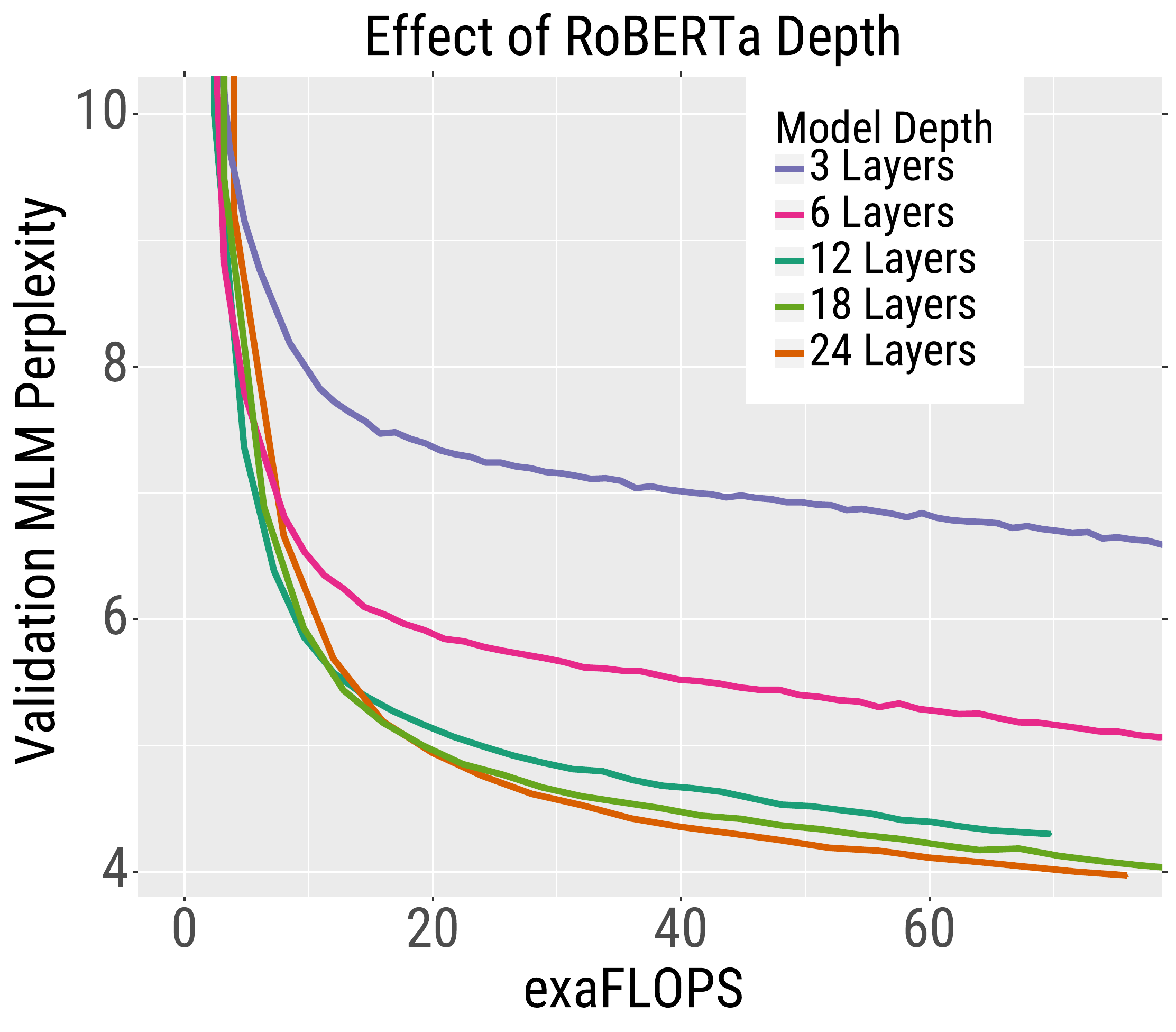}\hfill
\includegraphics[width=.33\textwidth]{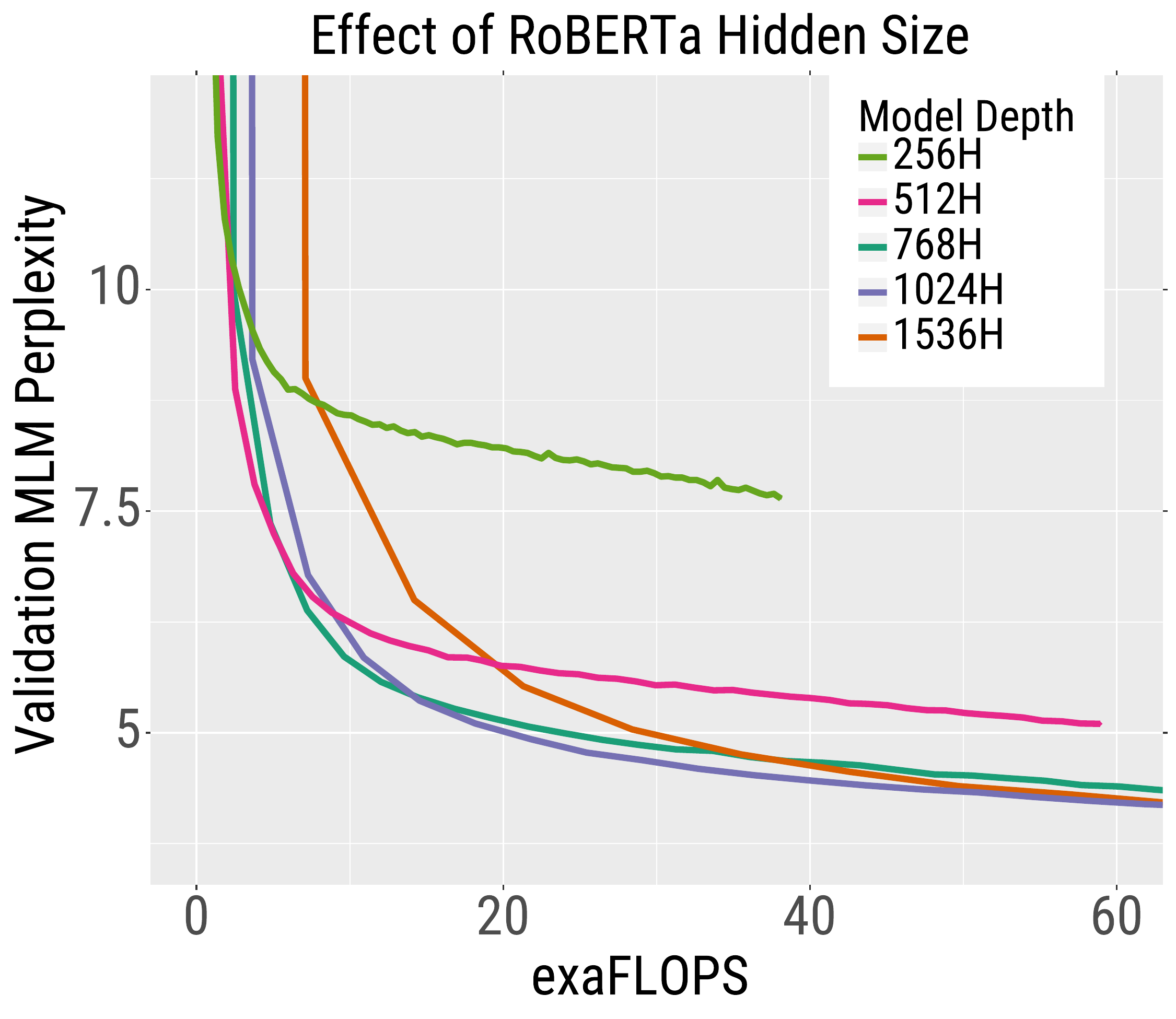}\hfill
\includegraphics[width=.318\textwidth]{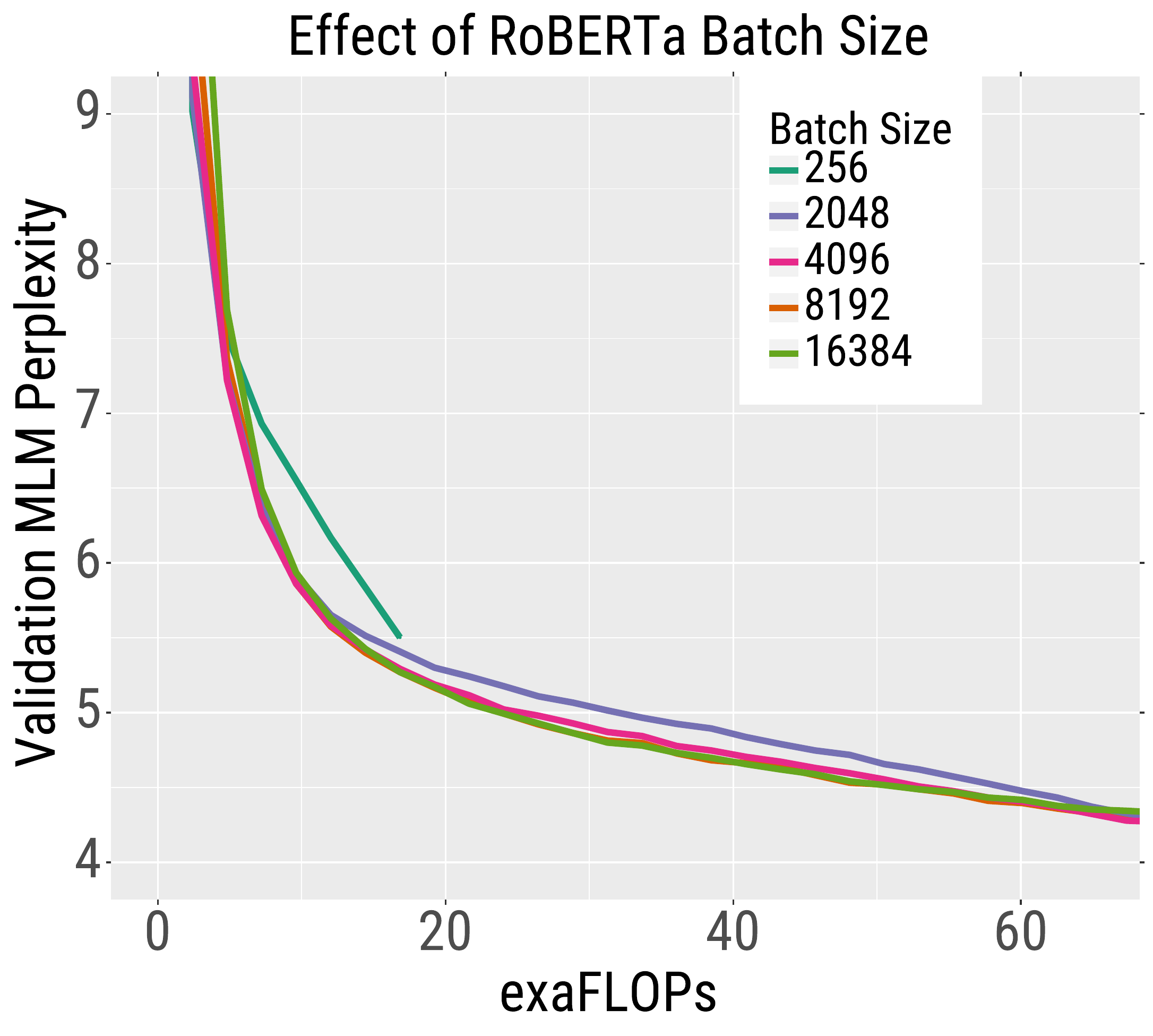}\hfill
\vspace{-0.3cm}
\caption{\emph{Floating Point Operations.} We show Figures~\ref{fig:intro}, \ref{fig:width}, and \ref{fig:batch} in terms of exaFLOPs instead of wall-clock time. Bigger models achieve better results than smaller models using the same number of floating point operations.} 
\label{fig:flops}
\end{figure*}

In Figure~\ref{fig:flops}, we plot selected learning curves from the main text as a function of FLOPs rather than seconds. We compute FLOPs using the code provided by \citet{clark2020electra}.

\subsection{The Impact of Batch Size}

\begin{figure*}[t]
\centering
\hfill
\includegraphics[width=.4\textwidth]{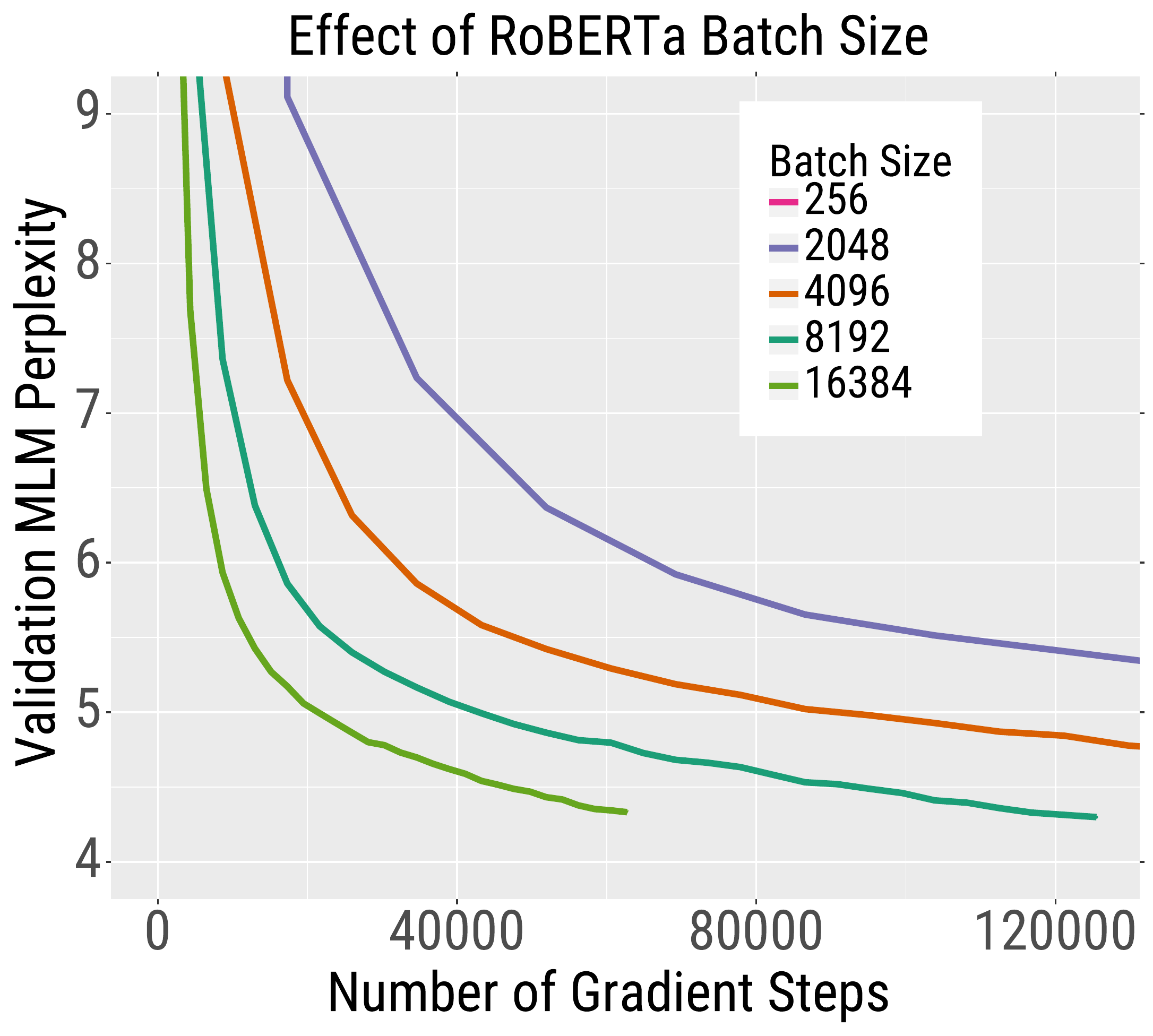}\hfill
\includegraphics[width=.415\textwidth]{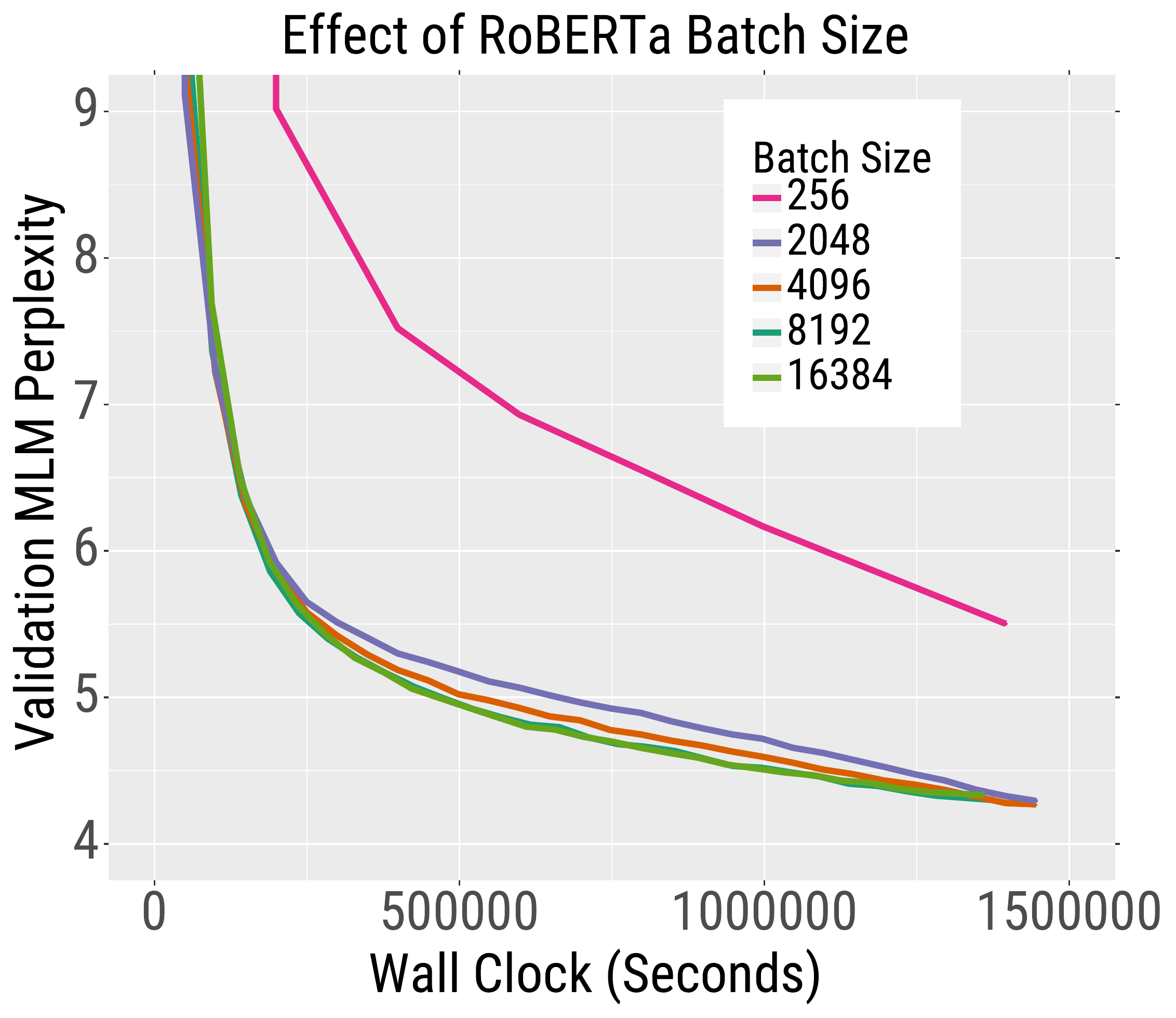}\hfill
\vspace{-0.3cm}
\caption{Increasing the batch size and the associated learning rate accelerates convergence in terms of gradient steps. However, increasing the batch size beyond 2048 provides only marginal improvements with respect to wall-clock time. Note that the wall-clock time includes the cost of accumulating gradients on a single machine (see Section~\ref{subsec:hardware}). In other words, beyond a certain point increasing the batch size only provides speedups when additional hardware is available. The 256 batch size result is far to the right in the left plot.}
\label{fig:batch}
\end{figure*}

Figure~\ref{fig:batch} shows the learning curves associated with different batch sizes. Table~\ref{table:learning_rates} shows the learning rates associated with each batch size. We use the hyperparameters from \citet{liu2019roberta} as a starting point and then lightly tune them.

\begin{table}[h]
\centering
\begin{tabular}{ll}
\toprule
\textbf{Batch Size} & \textbf{Learning Rate} \\
\midrule
256 &  .0002 \\
2048 & .001 \\
4096 & .00125 \\
8192 & .0015 \\
16384 & .001875\\
\bottomrule
\end{tabular}
\caption{The learning rate for each batch size in Figure~\ref{fig:batch}.}
\label{table:learning_rates}
\end{table}

\subsection{The Impact of Dataset Size}

\begin{figure*}[t]
\centering
\includegraphics[width=.4\textwidth]{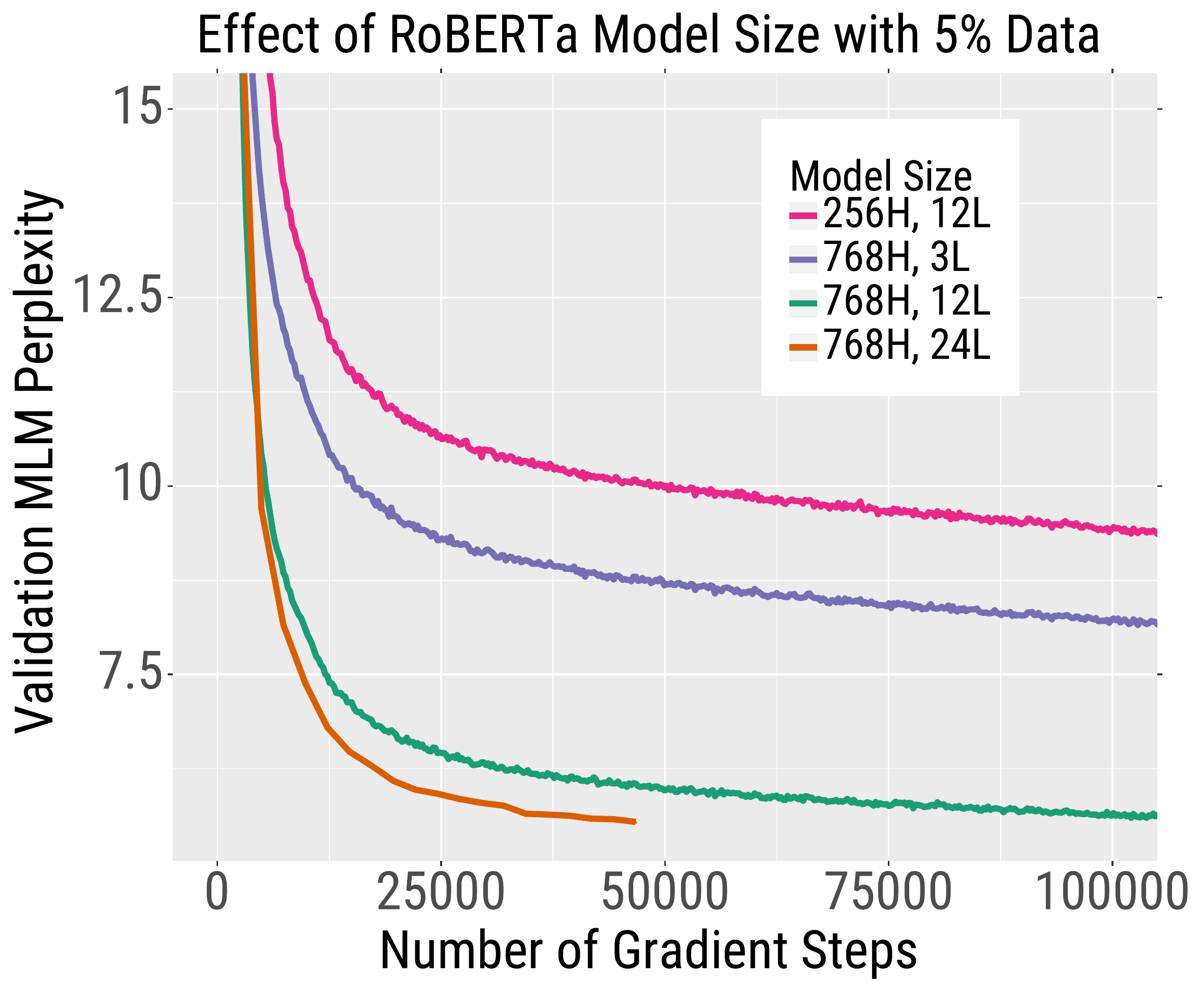}
\includegraphics[width=.4\textwidth]{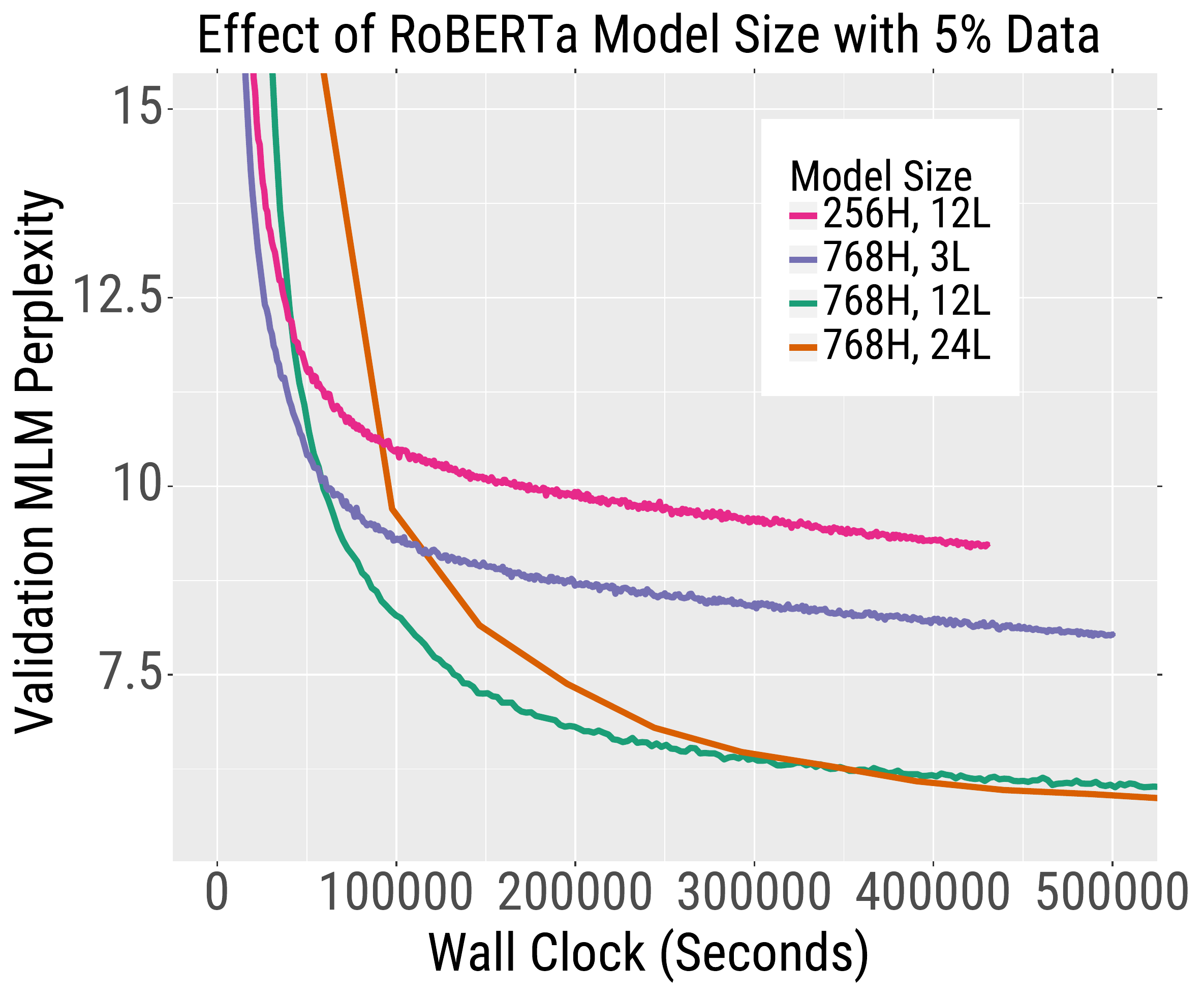}

\includegraphics[width=.4\textwidth]{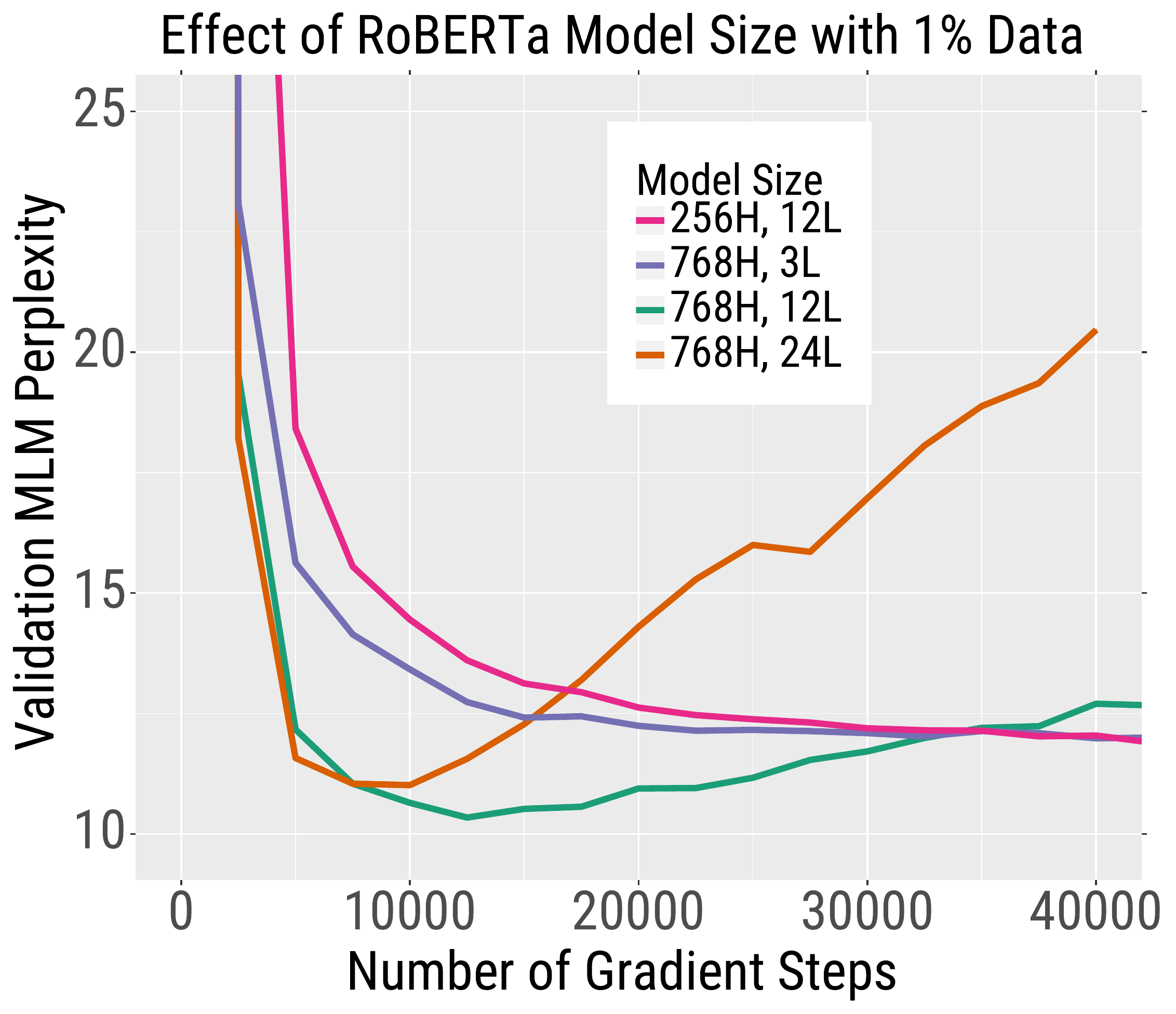}
\includegraphics[width=.39\textwidth]{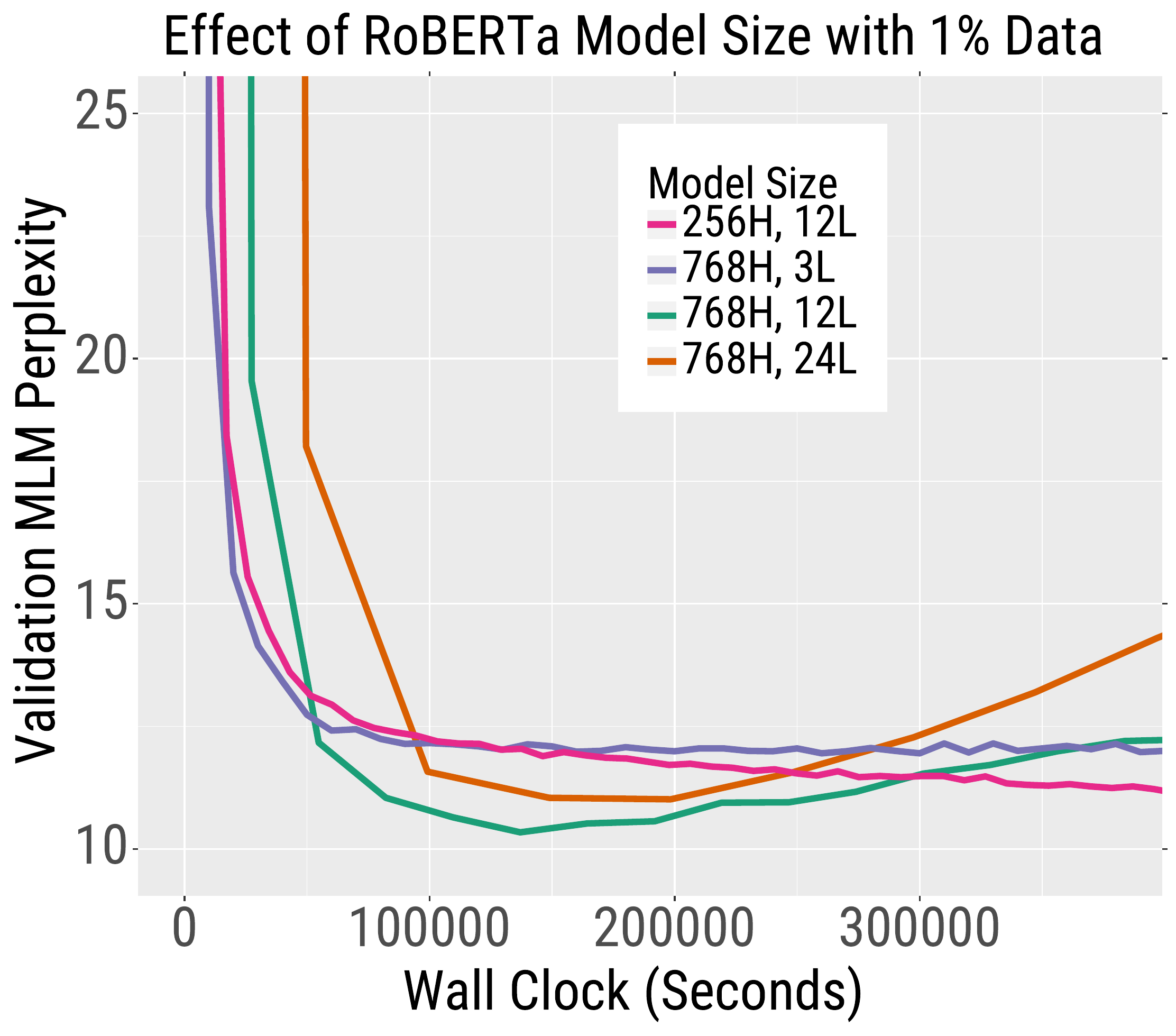}
\vspace{-0.3cm}
\caption{\emph{Effect of Smaller Datasets.} In our experiments on the full dataset (see main text), the largest models we trained are always faster in terms of wall-clock time. However, when subsampling the data to 5\% (top row), the biggest models do not improve on the speed of the smaller models (e.g., compare 24 Layer \roberta{} and 12 Layer \roberta{}). When the data is subsampled to 1\% (bottom row), the bigger models are \emph{worse} in terms of perplexity due to overfitting. This illustrates that the optimal model size depends on the dataset size.}
\label{fig:subsample_data}
\end{figure*}

Figure~\ref{fig:subsample_data} shows the learning curves for models trained using 5\% and 1\% of the training data.

\section{Finetuning Models of Different Sizes}\label{appendix:bigger_finetuning}

Table~\ref{table:bigger_finetuning} shows that models with more parameters are not harder to finetune.

\begin{table}[h]
\centering
\begin{tabular}{cccc}
\toprule
\textbf{Model} & \textbf{Perplexity} & \textbf{MNLI} & \textbf{SST-2} \\
\midrule
12-layer, 768H & 4.3 & 84.3 & 93.0 \\
18-layer, 768H & 4.1 & 85.4 & 92.6 \\
24-layer, 768H & 4.0 & 85.2 & 93.1 \\

\addlinespace

12-layer, 768H & 4.3 & 84.3 & 93.0 \\
12-layer, 1024H & 3.9 & 85.5 & 93.2 \\
12-layer, 1536H & 4.3 & 85.1 & 93.8 \\
\bottomrule
\end{tabular}
\caption{We train \roberta{} models of different sizes and stop them at roughly the same pretraining perplexity (the bigger models are trained for less wall-clock time). We then finetune each model on MNLI and SST-2. All models reach comparable accuracies (in fact, the big models often outperform small ones), which shows that larger models are not harder to finetune.}
\label{table:bigger_finetuning}
\end{table}

\section{Negative Results: Layer Sharing}\label{appendix:sharing}

Sharing weights across transformer layers can provide a small or negligible degradation in final performance~\cite{lan2019albert,dehghani2018universal} while providing a reduction in memory consumption. In addition, models with shared layers are slightly faster to execute because they require less memory movement and reduced inter-device communication. Similar to \citet{lan2019albert}, we experiment with two types of layer sharing: sharing all layers and sharing only the attention layers.

\begin{figure}[H]
\centering
\includegraphics[trim={0cm 0.5cm 0cm 0.0cm},clip, width=0.9\columnwidth]{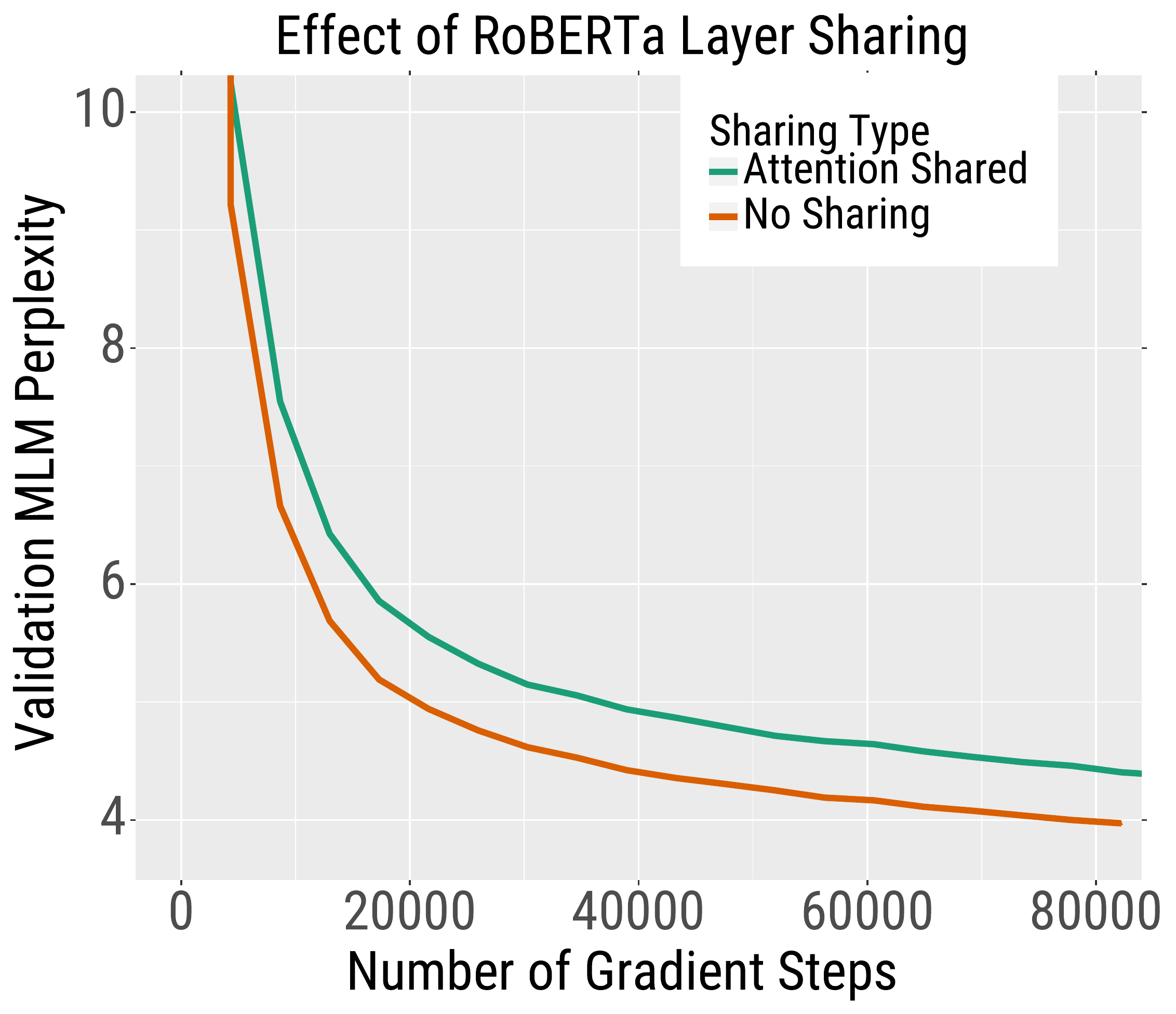}
\caption{Sharing attention layers reduces the maximum memory consumption of \roberta{} but causes slower convergence and worse final accuracy.}
\label{fig:sharing}
\end{figure}

Sharing layers reduces the maximum memory requirements, especially for small batch sizes. For example, sharing all the layers of a \roberta{} model with batch size 32 reduces total memory usage by 41\%. However, both forms of sharing lead to slower training convergence and thus worse performance in the resource-constrained setting (Figure~\ref{fig:sharing}). Consequently, we do not recommend sharing layers for compute-efficient training or inference of transformers.

\section{Compression Results for SST-2}\label{appendix:compression}

\begin{figure*}[t]
\centering
\hfill
\includegraphics[width=.32\textwidth]{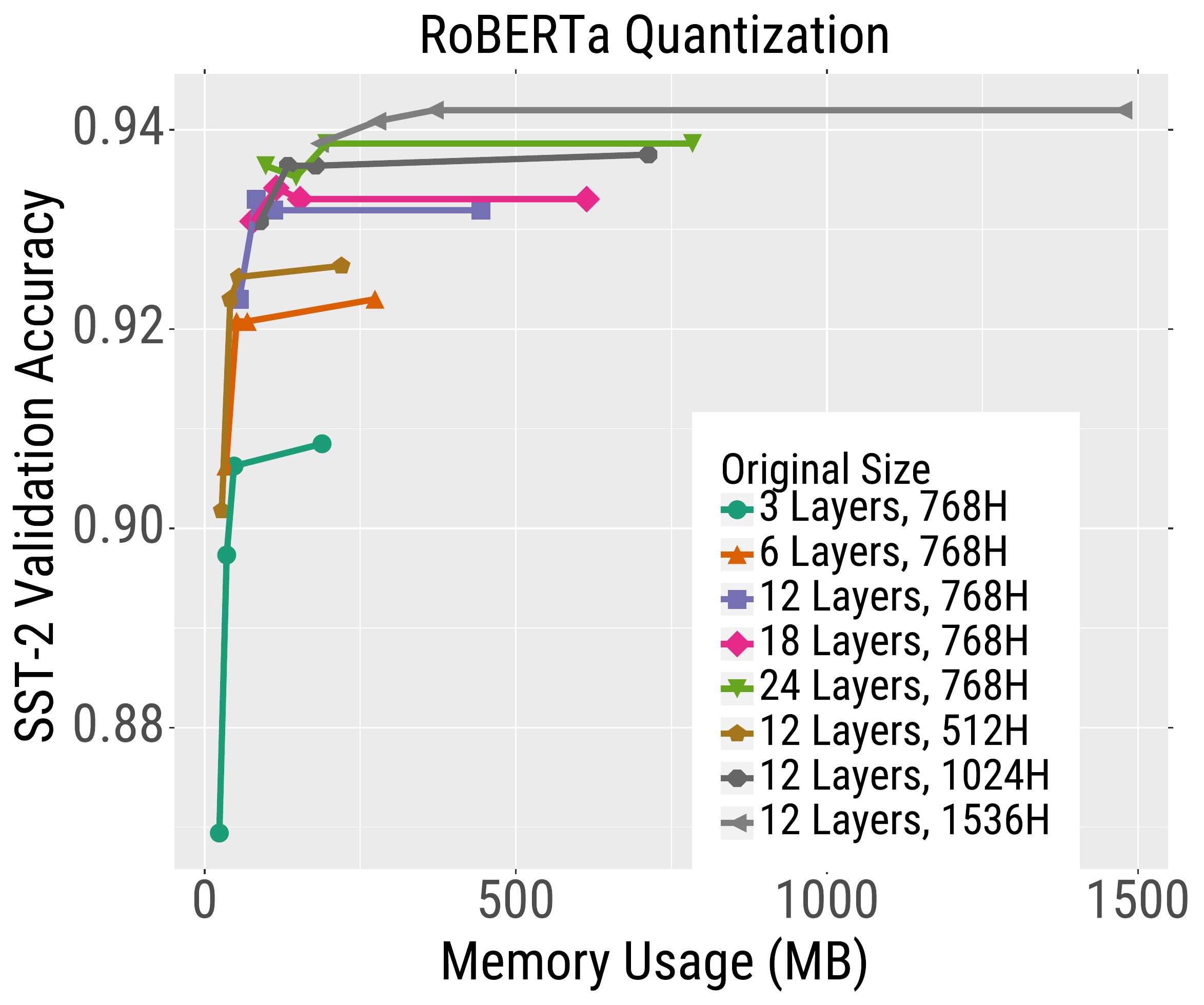}\hfill
\includegraphics[width=.32\textwidth]{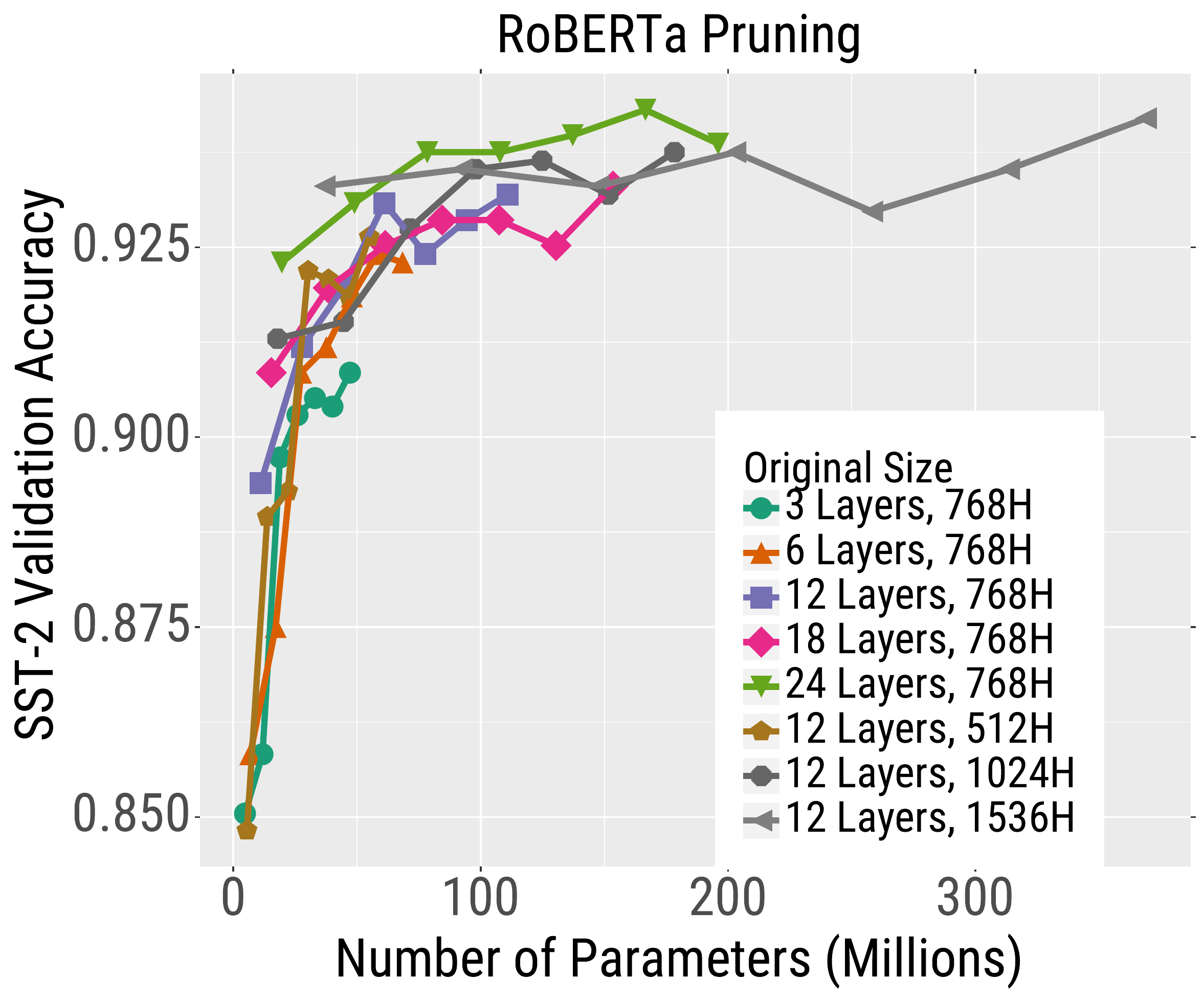}\hfill
\includegraphics[width=.32\textwidth]{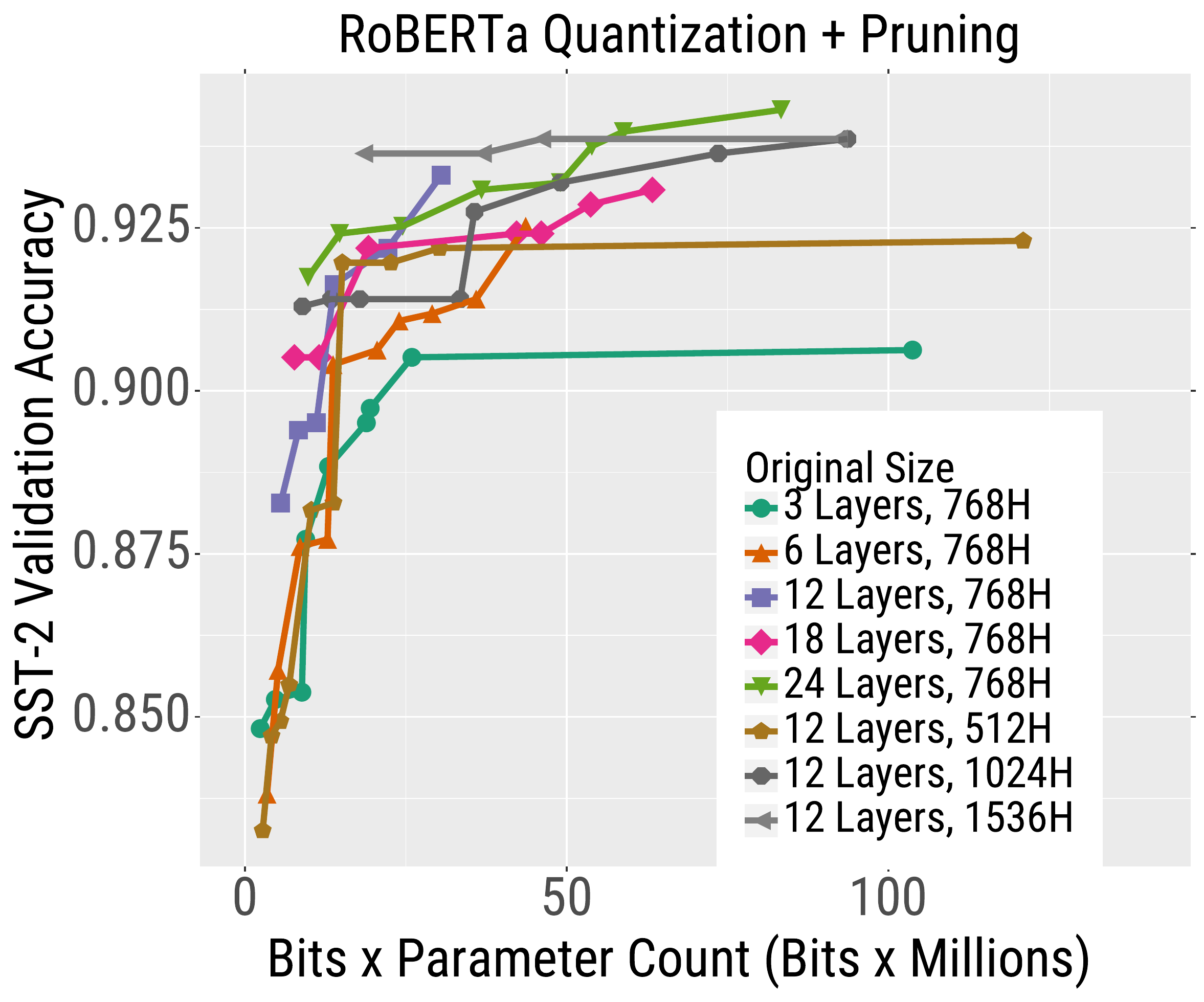}\hfill
\vspace{-0.3cm}
\caption{\emph{Compression for SST-2.} For most budgets (x-axis), the highest accuracy SST-2 models are the ones which are trained large and then heavily compressed. We show results for quantization (left), pruning (center), and quantization and pruning (right).}
\label{fig:sst}
\end{figure*}

We follow \citet{liu2019roberta} and report results on SST-2~\cite{socher2013recursive} in addition to MNLI. Since the SST-2 dataset is smaller than MNLI it requires a more significant tuning of the finetuning hyperparameters. We tune the batch size in $\{16, 32, 64\}$, the learning rate in $\{5e-4, 3e-4, 1e-4\}$, the seed which controls the classifier initialization and training data shuffling in $\{100, 300, 500\}$, and the dropout in $\{0.1, 0.2, 0.3\}$. We choose the best value using the validation set for each model size. We then perform quantization, pruning, and quantization and pruning on all finetuned models. Similar to MNLI, the bigger models provide the highest accuracy for a given test budget (Figure~\ref{fig:sst}).

\end{document}